\documentclass[12pt,a4paper,twoside]{article}

\makeatletter\if@twocolumn\PassOptionsToPackage{switch}{lineno}\else\fi\makeatother

\usepackage{amsfonts,amssymb,amsbsy,latexsym,amsmath,tabulary,graphicx,times,xcolor}
\usepackage[T1]{fontenc}
\usepackage{url,multirow,morefloats,floatflt,cancel,tfrupee,multicol}
\usepackage{ulem}
\usepackage{xcolor}

\newcommand{\add}[1]{\textcolor{black}{#1}}
\newcommand{\extraadd}[1]{\textcolor{black}{#1}}
\makeatletter

\AtBeginDocument{\@ifpackageloaded{textcomp}{}{\usepackage{textcomp}}}
\makeatother
\usepackage{colortbl}
\usepackage{xcolor}
\usepackage{pifont}
\usepackage[nointegrals]{wasysym}
\usepackage[superscript,biblabel]{cite}
\makeatletter
\usepackage{comment}
\usepackage{bm}
\usepackage{amsmath}
\usepackage{amssymb}
\usepackage{makecell}

\usepackage{booktabs}

\@ifundefined{etal}{}{}

\AtBeginDocument{
\expandafter\ifx\csname eqalign\endcsname\relax
\def\eqalign#1{\null\vcenter{\def\\{\cr}\openup\jot\m@th
  \ialign{\strut$\displaystyle{##}$\hfil&$\displaystyle{{}##}$\hfil
      \crcr#1\crcr}}\,}
\fi
}

\newcommand{\equref}[1]{(\ref{#1})}
\newcommand{\figref}[1]{Fig.~\ref{#1}}
\newcommand{\tabref}[1]{Table.~\ref{#1}}
\newcommand{\secref}[1]{Sec.~\ref{#1}}
\newcommand{\appref}[1]{Appendix.~\ref{#1}}

\@ifundefined{tsGraphicsScaleX}{\gdef\tsGraphicsScaleX{1}}{}
\@ifundefined{tsGraphicsScaleY}{\gdef\tsGraphicsScaleY{.9}}{}
\def\checkGraphicsWidth{\ifdim\Gin@nat@width>\linewidth
	\tsGraphicsScaleX\linewidth\else\Gin@nat@width\fi}

\def\checkGraphicsHeight{\ifdim\Gin@nat@height>.9\textheight
	\tsGraphicsScaleY\textheight\else\Gin@nat@height\fi}

\def\fixFloatSize#1{}
\let\ts@includegraphics\includegraphics

\def\inlinegraphic[#1]#2{{\edef\@tempa{#1}\edef\baseline@shift{\ifx\@tempa\@empty0\else#1\fi}\edef\tempZ{\the\numexpr(\numexpr(\baseline@shift*\f@size/100))}\protect\raisebox{\tempZ pt}{\ts@includegraphics{#2}}}}

\AtBeginDocument{\def\includegraphics{\@ifnextchar[{\ts@includegraphics}{\ts@includegraphics[width=\checkGraphicsWidth,height=\checkGraphicsHeight,keepaspectratio]}}}

\DeclareMathAlphabet{\mathpzc}{OT1}{pzc}{m}{it}

\def\URL#1#2{\@ifundefined{href}{#2}{\href{#1}{#2}}}

\edef\fntEncoding{\f@encoding}

\makeatother

\newif\ifmultipleabstract\multipleabstractfalse%
\usepackage[hmargin=1.8cm,vmargin=2.2cm]{geometry}
\makeatletter
\def\author#1{\gdef\@author{\hskip-\dimexpr(\tabcolsep)\hskip1pt\parbox{\dimexpr\textwidth-1pt}{\centering #1}}}

\let\@articletype\@empty \def\articletype#1{\gdef\@articletype{{\fontsize{14}{16}\selectfont #1}}}

\providecommand{\RunningHead}{}    
\providecommand{\RunningAuthor}{}  

\usepackage{fancyhdr}

\fancypagestyle{headings}{\fancyhf{}\fancyhead[RE]{\MakeTextUppercase{\textbf{\RunningAuthor}}}\fancyhead[LE]{\textbf{\thepage}}\fancyhead[LO]{\MakeTextUppercase{\textbf{\RunningHead}}}\fancyhead[RO]{\textbf{\thepage}}}
\fancypagestyle{plain}{\fancyhf{}\fancyfoot[C]{\textbf{\thepage}}}

\AtBeginDocument{\usepackage{lastpage}}
\def\title#1{%
  \gdef\@title{%
    \ifx\@articletype\@empty\else\@articletype~\\\fi%
     #1}%
}

\usepackage{abstract}
\def\abstractname{\textbf{Abstract}}
\renewenvironment{onecolabstract}
{\vspace*{-.4pc}\trivlist\item[]\leftskip1pt\noindent\selectfont\hfill\abstractname\hfill\mbox{\null}\par\ignorespaces}{\endtrivlist}

\def\NormalBaseline{\def\baselinestretch{1.1}}

\usepackage{textcase}
\usepackage[explicit]{titlesec}
\titleformat{\section}[block]{\NormalBaseline\boldmath\bfseries}
{\thesection.}
{6pt}
{#1}
[]
\titleformat{\subsection}[hang]{\NormalBaseline\filright\itshape}
{\thesubsection.}
{6pt}
{#1}
[]
\titleformat{\subsubsection}[block]{\NormalBaseline\filright\itshape}
{\thesubsubsection}
{6pt}
{#1}
[]
\titleformat{\paragraph}[runin]{\NormalBaseline}
{\theparagraph}
{6pt}
{#1}
[]
\titleformat{\subparagraph}[runin]{\NormalBaseline}
{\thesubparagraph}
{6pt}
{#1}
[]

\titlespacing{\section}{0pt}{1.5\baselineskip}{.2\baselineskip}  
\titlespacing{\subsection}{0pt}{1.5\baselineskip}{.2\baselineskip}  
\titlespacing{\subsubsection}{0pt}{1.5\baselineskip}{.2\baselineskip}  
\titlespacing{\paragraph}{0pt}{.5\baselineskip}{10pt}  
\titlespacing{\subparagraph}{0pt}{.5\baselineskip}{10pt}

\usepackage{caption}
\DeclareCaptionLabelFormat{figlabel}{Figure #2}
\date{}
\makeatother

\usepackage{endfloat}
\usepackage{float}
\usepackage{bm}
\usepackage{siunitx}

\newif\ifdraft
\draftfalse 
\newcommand{\revise}[1]{\textcolor{\ifdraft blue\else black\fi}{#1}}

\begin{document}

\title{Flexible Morphing Aerial Robot with Inflatable Structure for Perching-based Human-Robot Interaction}

\author{Ayano Miyamichi$^{1}$, Moju Zhao$^{2}$, \add{Kazuki Sugihara$^{1}$, Junichiro Sugihara$^{1}$, Masanori Konishi$^{1}$,}  Kunio Kojima$^{1}$, Kei Okada$^{1}$, Masayuki Inaba$^{1}
\thanks{$^{1}$Department of Mechano-Infomatics, The University of
Tokyo $^{2}$Department of Mechanical Engineering, The University
of Tokyo, 7-3-1 Hongo, Bunkyo-ku, Tokyo 113-8656, Japan.
        {\tt\small miyamichi@jsk.imi.i.u-tokyo.ac.jp}%
}$
}
\maketitle


{\begin{onecolabstract}
Birds in nature perform perching not only for rest but also for interaction with human such as the relationship with falconers.
Recently, researchers achieve perching-capable aerial robots as a way to save energy, and deformable structure demonstrate significant advantages in efficiency of perching and compactness of configuration. \add{However, ensuring flight stability remains challenging for deformable aerial robots due to the difficulty of controlling flexible arms.} 
Furthermore, perching for human interaction requires \add{high compliance} along with safety.
Thus, this study aims to develop a \add{deformable} aerial robot capable of perching on humans with high flexibility and \add{grasping} ability.
To overcome the challenges of stability of both flight and perching, we propose a hybrid morphing structure that combines a unilateral flexible arm and a pneumatic inflatable \add{actuators}. This design allows the robot's arms to remain rigid during flight and soft while perching for more effective \add{grasping}.
We also develop a pneumatic control system that optimizes pressure regulation while integrating shock absorption and adjustable grasping forces, enhancing interaction capabilities and energy efficiency.
\extraadd{Besides, we focus on the structural characteristics of the unilateral flexible arm and identify sufficient conditions under which standard quadrotor modeling and control remain effective in terms of flight stability.}
Finally, the developed prototype demonstrates the feasibility of \add{compliant} perching maneuvers on humans, as well as the robust recovery even after \add{arm deformation caused by thrust reductions during flight}.
To the best of our knowledge, this work is the first to achieve an aerial robot capable of perching on humans for interaction.

\def\keywordstitle{Keywords}
\end{onecolabstract}}
 
\begin{multicols}{2}
\section{INTRODUCTION} \label{sec1}
Aerial robots have been extensively utilized for tasks such as surveillance \cite{Bonatti2020}, inspection \cite{Sewer}, and disaster relief \cite{Michael2012} in locations inaccessible to humans, owing to their high mobility \cite{Floreano2015, kumar2012}.
However, \revise{the practicality of aerial robots in most of the application is still limited due to the significantly higher energy consumption compared to other types of robots.}
Thus, some \revise{researchers have} focused on perching \revise{ability for} aerial robots, inspired by birds to rest perching on trees.
Perching, defined as landing and grasping onto objects \cite{hang2019perching}, allows aerial robots to save energy and acquire high-quality data from \revise{stationary positions}\cite{danko2005robotic}.
\add{However, these aerial robots are mainly active in areas far from humans, and opportunities to utilize their mobility in human living spaces are limited. In human living spaces, several researches use aerial robots to provide haptic or visual guidance to impaired users\cite{allenspach2022human,liao2021robotic}, while rapid vital checks and rescue triage are also a potential use for aerial robots \cite{jain2018comparison,sanz2022drone}.}
While current aerial robots can only perch on static and arboreal objects \add{and current aerial robots for interaction are often active in the air, facing challenges to establish closer interactions and provide direct support}, birds \revise{in nature}, such as falconer-trained hawks and pet birds, \revise{can come into} physical contact with humans by perching on the shoulders or arms as a means of not only rest, but also bonding and communication\cite{anderson2003bird,anderson2014social}.
\add{By applying bird behavior to aerial robots, through perching-based physical contact to human, aerial robots can establish a more intimate relationship and acquire and provide more direct and reliable data to humans.
Specifically, in disaster rescue scenarios, a perching robot could acquire more accurate biometric information for rapid vital checks and triage or provide haptic guidance to stranded individuals.}

\begin{figure}[h!]
    \centering
    \includegraphics[width=1.0\columnwidth]{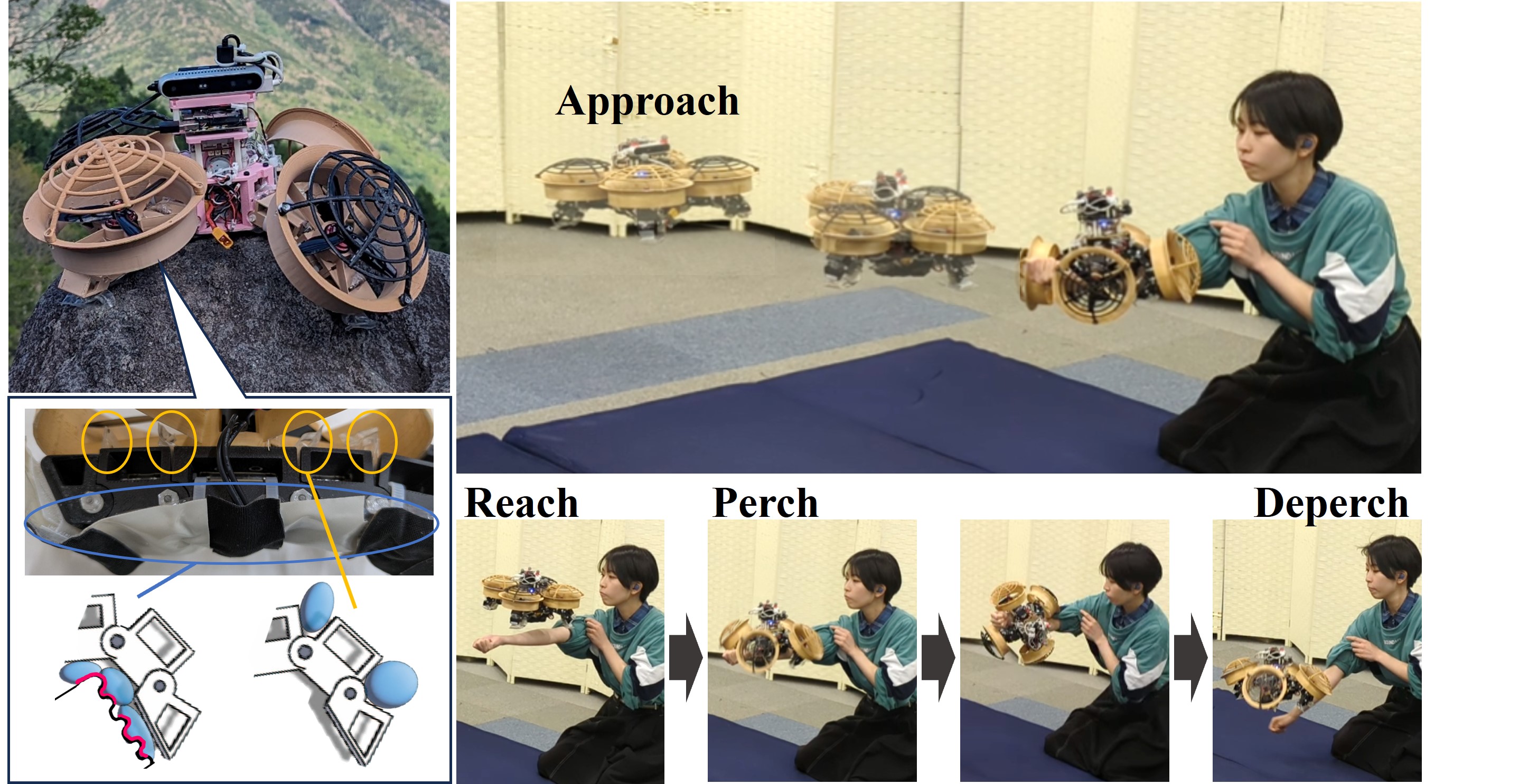}
    \caption{\color{black}Proposed morphing aerial robot composed of unilateral flexible arms and a pneumatic inflatable mechanism, which can perch on human arm dynamically in in-flight situation}
    \label{intro}
\end{figure}

Therefore, inspired by these birds interacting with humans, this work aims to pioneer a new aerial robot platform that can perch on human body parts to perform new human-aerial robot interactions\cite{tezza2019state} through perching \revise{as depicted in \figref{intro}}.

\subsection{Related Works}
Current aerial robots \add{demonstrate various approaches to achieve perching, which} can be classified into two main categories based on the body configuration: those equipped with independent grasping modules \add{\cite{mclaren2019passive,5980314,lin2024ultra,8793936,zhao2024design,roderick2021bird, ubellacker2024high,doyle2012avian,nguyen2023soft,9828332,pingaerial,10521918}} and those with deformable bodies\add{\cite{bucki2022design,tao2023design,9851515,zheng2023metamorphic}}.
The configurations with an independent grasping mechanism at the bottom of the robot guarantee a relatively high stability by decoupling flight and grasping function.
However the overall weight and size increase.
On the other hand, deformable aerial robots \revise{embed rotors into the morphing arms, which leads to the integration of flight and grasping functions for more efficient perching using their entire body and thus minimizing size and weight\cite{tilting_frame_quad,figure_8,morphing_quad,n_zhao,origami,Bucki,hydrus,Lasdra}.}
\add{In this configuration, there is a trade-off between flight stability and adaptability. For example, rigid arms with hinges \cite{bucki2022design,tao2023design}helps maintain stability during flight but makes the robot less adaptable when perching on objects.
In contrast, soft arms\cite{9851515,zheng2023metamorphic} improve adaptability for perching but negatively affect flight stability due to unexpected changes in the rotor configuration.}

In addition, the high compliance of adaptive grasping should be considered crucial when perching on the human body. \add{Recent advancements in soft robotics have yielded pneumatically driven flexible structures that offer promise for this application. 
These structures enhance both impact absorption and interaction capabilities by dynamically controlling grasping force and shape through air pressure modulation. However, most studies on inflatable actuators focus on improving grasping\cite{9828332,pingaerial,10521918} and impact absorption\cite{cawthorne2016development, ansari2023design,nguyen2023soft} as separate functions, which results in inefficient energy use and limited functionality. 
For example, pneumatically actuated silicone graspers\cite{9828332,pingaerial,10521918} are widely used for soft and adaptive grasping but often face challenges in load-bearing capacity and grasping speed. 
Conversely, Nguyen et al. \cite{nguyen2023soft} developed an inflatable frame to absorb landing impact, coupled with a passive grasper that flexes by embedded leaf springs and is recoiled by air pressure. However, in this design, the function of the pneumatic system is limited to shock absorption.  
While passive grasping mechanisms address the limitations of soft graspers, the grasping force cannot be adjusted, which could lead to discomfort or injury during human interaction.}

Recent studies on deformable perching aerial robots have focused on incorporating deformation and nonlinear dynamics into their modeling and control to improve adaptability\cite{9851515,ryll2022smors}. For example, Ryll et al.\cite{ryll2022smors} present a system that combines soft robotics and aerial capabilities. However, these approaches often require intricate designs and computations, making them less practical for practical  applications. 
Moreover, the stability of perching on the target object in in-flight situation is still challenging.

In order to overcome the challenges faced by both configurations, we introduce a deformable aerial robot with rotor-embedded arms that combine flexibility for compliant perching on humans and rigidity for flight stability. 
This hybrid design, which merges rigid and soft elements, enables the robot to generate greater grasping force than arms made solely of soft inflatable materials or rigid structures.
Furthermore, this integrated design allows the robot to become compact while perching on human body, which enhances safety and wearability.
Given that the fabrication method is the key to the inflatable module \cite{zhou2022direct, bruton2016packing}, in this work an original design is proposed to improve the controllability of the flexible arms that use inflatable actuators for shock absorption and controllable grasping.
These features allow the robot to interact safely with the human body while providing soft and adjustable perching forces to reduce discomfort and the risk of injury.
Besides, the proposed pneumatic system combine both functions unlike existing designs and contributes to quick deformation of the robot arm and efficient energy use.
In addition, a modeling and control framework for the proposed morphing aerial robot composed of unilateral flexible arms is also necessary to achieve bistability during flight and grasping.
We focus on the unique characteristics of unilateral flexible arms, which allows us to approximate a much simpler dynamics mode around the hovering situation.
\extraadd{By} assuming that the impact of arm deformation during flight is negligible, \extraadd{standard quadrotor control can be effectively applied to the morphing aerial robot, thereby achieving robust flight stability without requiring complex adaptation or state estimation.}

\subsection{Contribution}
The contributions of this study are outlined as follows:
\begin{itemize}
    \item We present a design methodology of a novel deformable aerial robot that combines unilateral flexible rotor-embedded arms with inflatable actuators for compliant perching ability.
    \item We develop a pneumatic actuation \extraadd{control} system specifically designed for perching, which integrates shock absorption and controllable grasping capabilities.
    \item \extraadd{We identify sufficient thrust condition that enables standard quadrotor-based control to achieve stable and robust flight, by leveraging the deformable arm’s structural characteristics.}
    \item We demonstrate the feasibility of autonomous and stable perching motion on the human body by the prototype robot, incorporating task planning for human-robot interaction through perching, along with robustness in flight and stable clinging during grasping.

\end{itemize}

\subsection{Organization}
\revise{
The paper is structured as follows: \secref{design} presents
the mechanical design of the proposed deformable aerial robot. 
 \secref{Pneumatic System} explains the pneumatic actuation system used in the robot. 
 \secref{control} outlines the modeling and control framework.
\secref{experiments} discusses the thorough experiments with the developed prototype involving the evaluation of grasping performance, flight stability and in-flight perching motion. Conclusions and new lines of research are proposed in \secref{conclusion}.
}

\section{Mechanical Design} \label{design}
\subsection{\revise{Unilateral} Flexible Arm} 
\begin{figure*}[t]
    \centering
    \includegraphics[width=1.0\columnwidth]{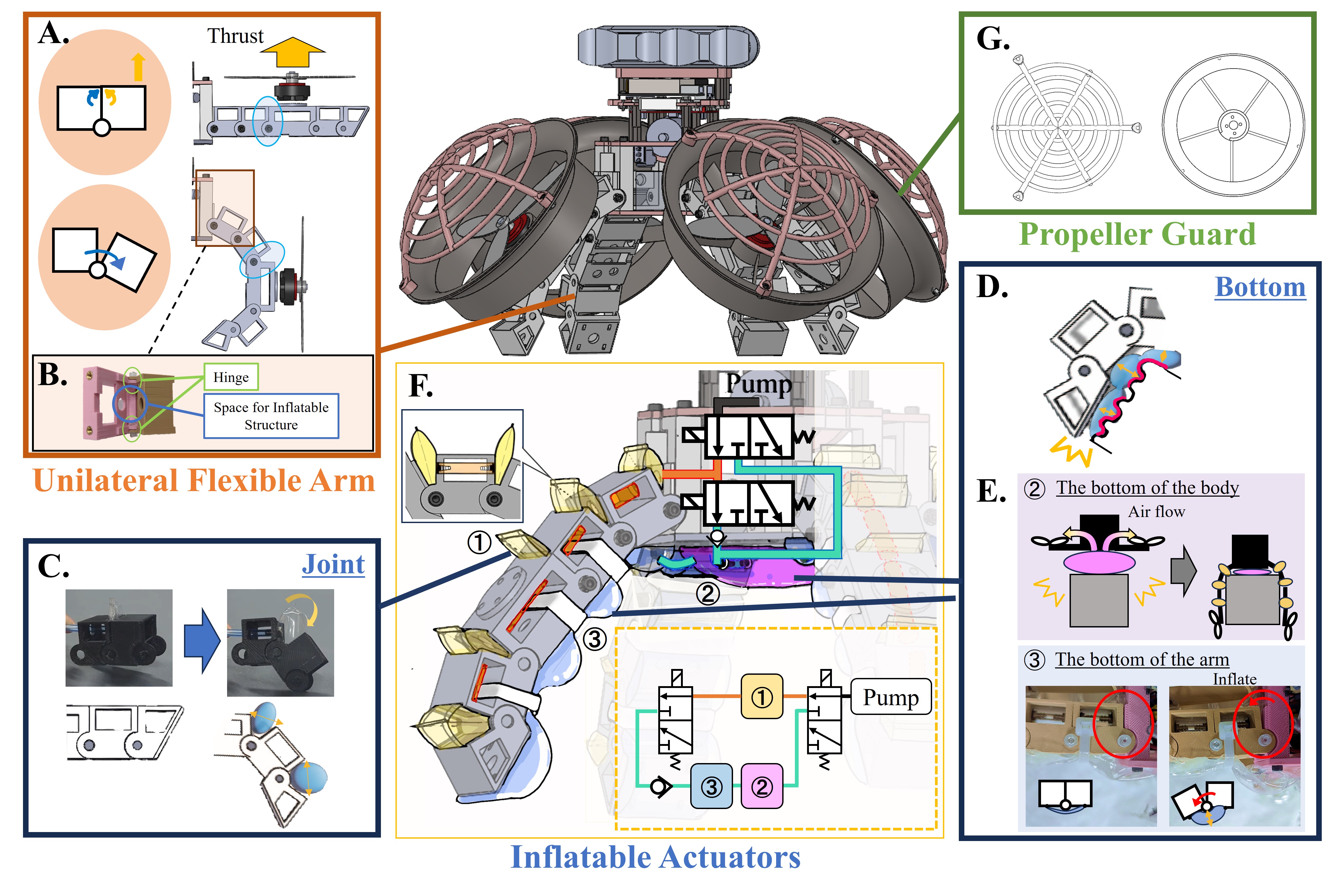}
    \caption{\revise{Proposed} aerial robot that can perch on human body.
      (A) \revise{design} of the unilateral flexible arm.
      (B) \revise{design} of the unilateral flexible arm. \revise{While} flying, thrust forces cause the arm to beam \revise{upward}, and when the robot is stationary, the arm \revise{deform downward}. The arm has hinges as joints and \revise{the small spaces for inserting inflatable actuators and tubes}.
      (C) \revise{function} of the inflatable mechanism for joints. \revise{It can move the related joint and thus deform the arm.}
      (D) \revise{main} function of the inflatable \revise{modules at the bottom}. They absorb the impact of landing and \revise{increase} the contact area with the \revise{target} object.
      (E) \revise{another} function of the inflatable \revise{module at the bottom}.
      The inflatable \revise{module} installed on the bottom of the body \revise{can} send additional air to the inflatable actuators \revise{embedded in arm joints as depicted in (C)}.
      \revise{The modules} on the bottom of the flexible arms can generate \revise{bending torque} to the joints at the root of the arm \revise{by expanding}.
      (F) \revise{air} flow path between inflatable actuators and pneumatic \revise{device}.
      (G) \revise{upper (left) and bottom (right) views of the propeller guard}.}
    \label{all_hardware_new}
\end{figure*}

To ensure bistability \revise{for} both flight and perching, the arm \revise{embedded} with the rotor should \revise{offer} rigidity for flight and flexibility for grasping during perching.
Thus, we develop a unilateral flexible arm capable of transitioning between these postures as shown in \figref{all_hardware_new}A.

The unilateral flexible arm comprises interconnected segments with a unidirectional, constrained range of motion. Thus, the arm can incorporate structural features that lock adjacent segments by bracing against each other as a rigid chain.
This locking mechanism prevents the arm from undergoing excessive deformation or torsion beyond expected thresholds.
Consequently, as depicted in \figref{all_hardware_new}B, \revise{those arms can perform} flexible deforming similar to a chain \revise{in idle condition}. On the other hand, during flight, when upward thrust by the propeller is applied to the arm, the locking mechanism ensures that each link supports the others, maintaining straight rigidity.

\add{As the arm length increases, the maximum graspable radius also increases, but interference between diagonal arms  limits the ability to grasp smaller diameter objects.}
\add{Therefore, the arm length is optimized to be as long as possible while still allowing the grasping of the smallest typical adult forearm without interference.
This design ensures adaptability to a wide range of forearm sizes across different individuals.
In addition, the rotor position is also optimized to compact the whole size and minimize interference between the ducted propeller guards during arm deformation.}

\add{Regarding the segment, each hollow arm segment is designed to accommodate inflatable actuators, tubing, and cables within its structure. Its length is minimized while maintaining structural strength and providing space for two connectors of joint inflatable actuators, allowing the placement of three hinges between the arm base and rotor.
Furthermore, maximum hinge angles are designed to prevent the arm from deforming beyond \( \pi \) \SI{}{rad} from the base of the arm to the rotor, ensuring that the thrust during flight can restore the arm attitude. Specifically, the base hinge allows for a maximum rotation of \( \frac{5\pi}{18} \) \SI{}{rad}, while the other hinges are designed to rotate up to \( \frac{\pi}{3} \) \SI{}{rad}.}

\subsection{Joint Inflatable Actuator for Grasping Force}          
In this study, we develop a pneumatic inflatable actuator for joints to actively deform the arm, as shown in \figref{all_hardware_new}C.
This inflatable actuator consists of an airbag with connectors and tubes. 
When pressurized, each airbag \revise{that is embedded in} the gap space between the arm segments expands and generates torque at the hinges, causing the arm to deform.
The air volume in
each airbag is automatically adjusted to conform to the shape of the grasped object, enabling appropriate joint angle adjustment and adaptive arm morphing.

\begin{figure}[b]
    \centering
    \includegraphics[width=0.8\columnwidth]{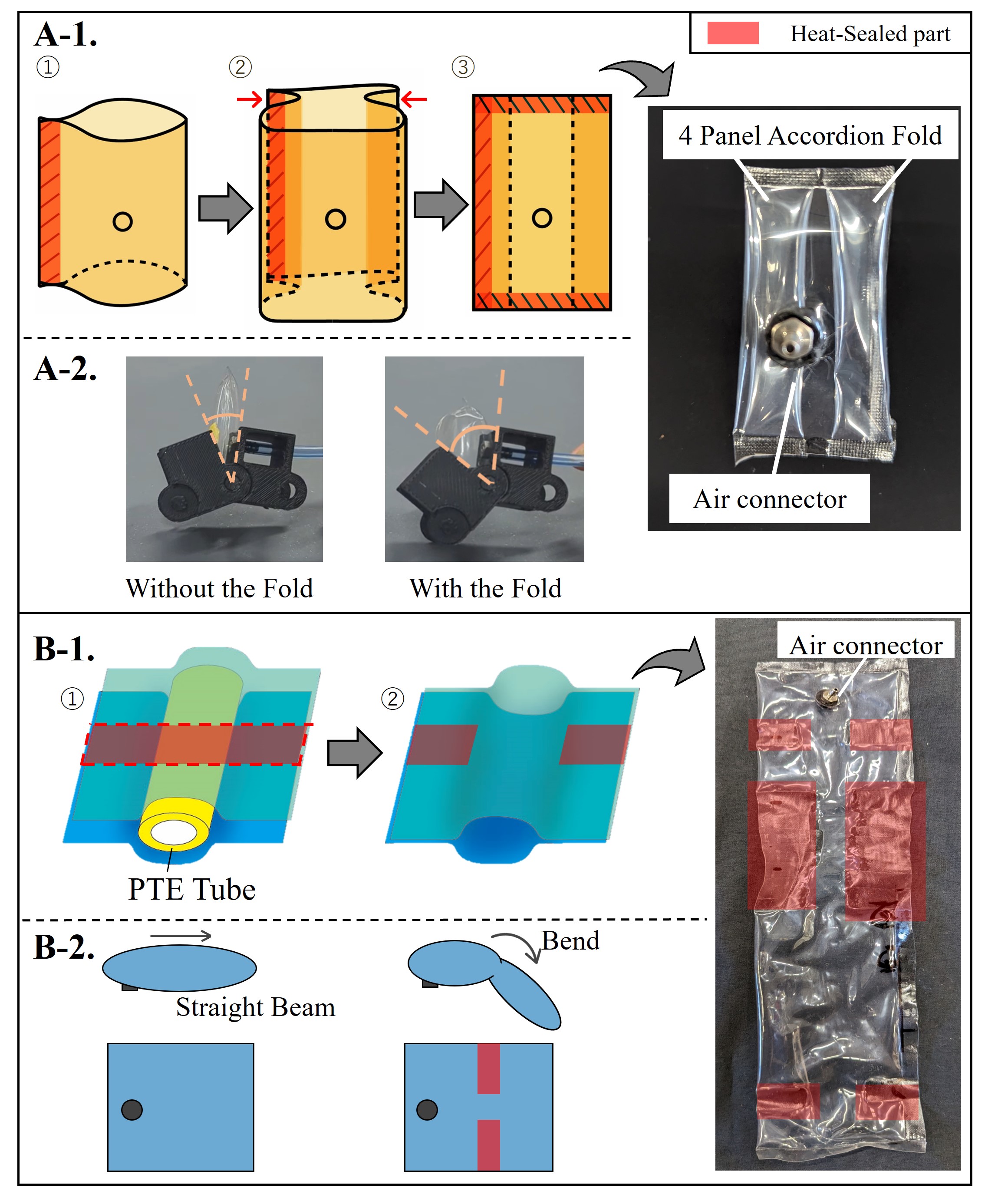}
    \caption{\revise{Fabrication process of the inflatable actuators for joints}. (A-1) inflatable actuator for joint fabrication. Both sides of the TPU are folded into four panel accordion fold and heat-sealed together.
      (A-2) \revise{left} figure shows the joint movement using the airbag without four panel accodion fold, and the right figure shows that using the airbag with the fold.
      (B-1) \revise{inflatable} actuator for the bottom of the arms fabrication.
      When making a joint, PTE Tube is inserted between the TPU sheets and heat-sealed as shown in the figure, which results in heat-sealing of the parts other than the tube.
      This allows for a larger flow path than when heat-sealing without the tube.
      (B-2) \revise{by} heat-sealing as shown in the figure on the right, the inflatable actuator can be articulated and flexed.}
    \label{inflatable}
\end{figure}

There are two challenges faced when designing these airbags.
\revise{One} is that the airbags, which inflate in the gaps between the segments, are prone to punctures due to contact or friction with other parts. To address this problem, each airbag is connected in series. It makes replacement easy and enhances modularity.
The second challenge is to maximize the airbag's volume to generate sufficient grasping force while compactly stowing the airbag between joints during flight.
To solve this issue, we develop a foldable airbag with both sides of the airbag with a four-panel accordion fold, as depicted in \figref{inflatable}A-1.
This design allows for increased air volume compared to conventional non-folded airbags, enabling the arm to deform to its structural limits, as illustrated in \figref{inflatable}A-2.

Therefore, by incorporating the joint inflatable actuator in the segment of the unilateral flexible arm, the airbags are wholly folded and stored between the links during flight. It allows the arm to behave as a rigid beam under thrust. 
Meanwhile, during grasping, the actuator expands via air supplied from the compact pump equipped at the center of the robot's body. This allows the arm to deform and perform compliant grasping with sufficient force.

\begin{figure}[h]
    \centering
    \includegraphics[width=1.0\columnwidth]{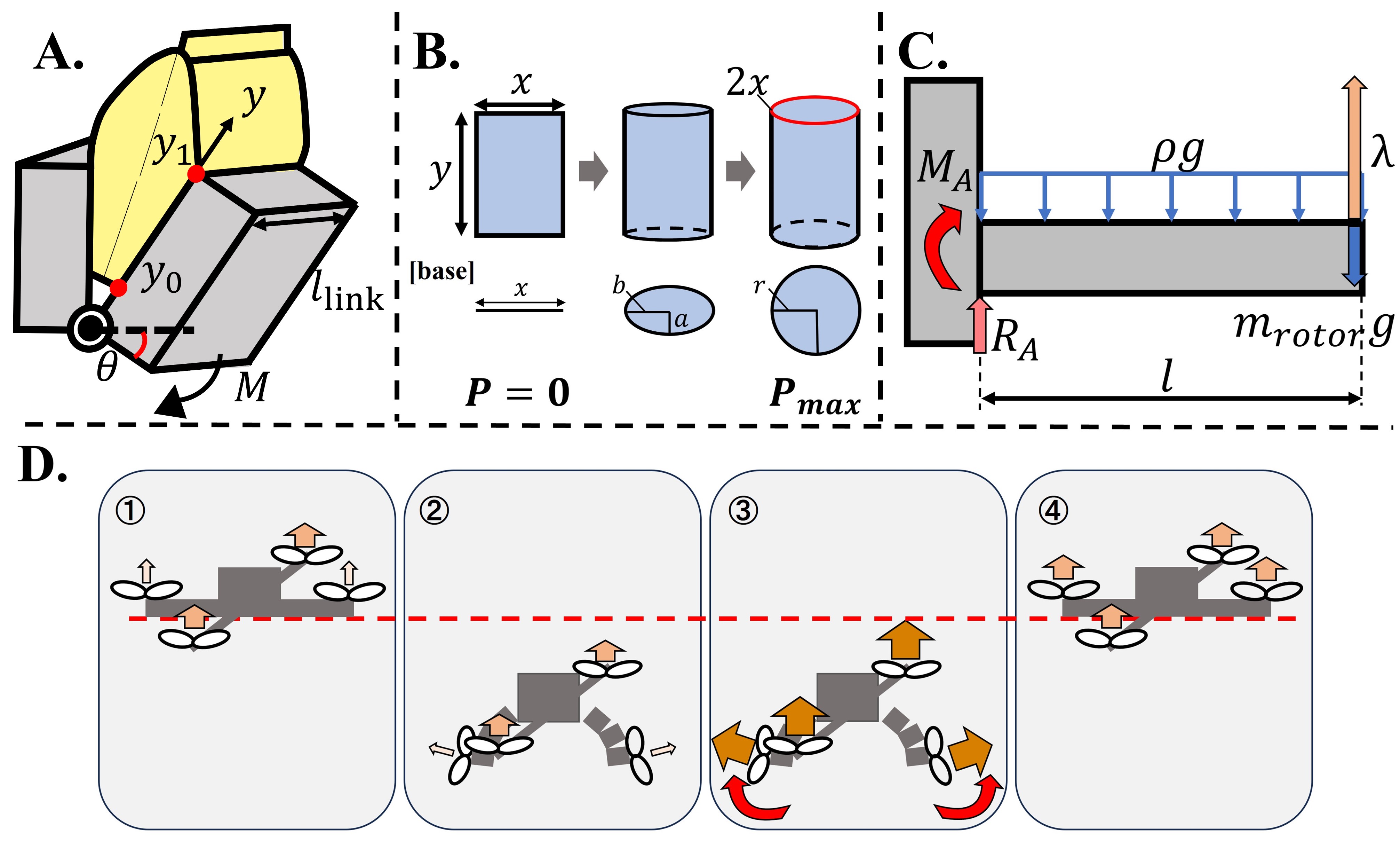}
    \caption{(A) The model of adding torque to a hinge by inflated a joint airbag in the unilateral flexible arm
    \add{(B)} the model of volume of a inflated airbag
    \add{(C) \revise{Rigid} model of the unilateral flexible arm}
    \add{(D)} \revise{desired recovery behavior} when the robot's arm deforms during flight.}
    \label{stable_order_new}
\end{figure}

To analyze how the joint inflatable actuator generates torque on the hinges through air pressure, we model the torque generation mechanism of a single joint airbag integrated into the unilateral flexible arm, as depicted in \extraadd{\figref{stable_order_new}A}.
Unless otherwise specified, all pressures in this paper are expressed as gauge pressure.
Assuming static fluid conditions, a linear elastic response of the material, and a pressure distribution dependent on the joint angle, the internal pressure distribution inside the airbag is modeled as follows:
\begin{equation}
    P(y) = k_0 P_0 - k_1 P_0 \theta- k_2x  \label{eq:Pa_1}
\end{equation}
where $P_0$ is the base internal gauge pressure inside the airbag when air is supplied from an external source, $k_0$ is a proportional coefficient, $k_1$ and $k_2$ are pressure gradient constants, $\theta$ is the hinge angle, and $x$ and $y$ are the horizontal and vertical distances from the hinge axis to the force application point, respectively.
Assuming that the pressure distribution depends primarily on the vertical displacement of the airbag, we approximate it as \( x \approx y \theta \), based on the small angle assumption.
Substituting this into \equref{eq:Pa_1} gives:
\begin{equation}
    P(y) \approx k_0 P_0 - k_1 P_0 \theta - k_2 y \theta
    \label{eq:Pa}
\end{equation}
The infinitesimal area of contact between the hinge and the airbag $ds$, can be expressed as:
\begin{equation}
ds = l_{\text{link}}dy
\end{equation}
where $l_{\text{link}}$ is the width of the link.
Then, the infinitesimal force $dF$ and infinitesimal torque $dM$ are given by:
\begin{equation}
dF = P(y) ds = P(y) l_{\text{link}} dy
\end{equation}
\begin{equation}
dM = ydF = P(y)l_{\text{link}} y dy
\end{equation}
Thus, the torque $M$ generated at the hinge is expressed as:
\begin{align}
M &= \int_{y_0}^{y_1} \left( k_0 P_0 - k_1 P_0 \theta - k_2y \theta \right) l_{\text{link}}  y dy \\
    &= \frac{1}{2} (k_0 - k_1 \theta) P_0 l_{\text{link}} \left( y_1^2 - y_0^2 \right) \notag \\ 
    &\quad- \frac{1}{3} k_2 \theta l_{\text{link}} \left( y_1^3 - y_0^3 \right) 
\label{eq:torque}
\end{align}
\equref{eq:torque} allows the torque on hinge to be expressed as a function of the joint angle $\theta$ and the pressure $P_0$ of joint airbag.
In this study, the values of $k_0$, $k_1$, and $k_2$ are determined experimentally, with the details provided in \appref{joint torque experiment}. 
The model effectively approximates the measured values, demonstrating its validity in capturing the overall trend of the torque generation mechanism of the joint inflatable actuator, though deviations are observed at higher pressures and larger angles.

\subsection{\revise{Bottom} Inflatable Actuator for Flexible Grasping} \label{bottom_design}
As shown in \figref{all_hardware_new}D, we also develop an inflatable actuator for the bottom of the robot, constantly expanding during operation.
This bottom inflatable actuator comprises airbags attached to the body's and arm's undersides, each connected by tubes.
Its main functions are absorbing the landing impact during perching to improve grasping flexibility and increasing the contact area with the object to absorb landing errors and improve grasping stability.
Additionally, as illustrated in \figref{all_hardware_new}E, this actuator has two auxiliary functions for the unilateral flexible arms.

Firstly, the airbag installed on the underside of the body \revise{can act} as an air tank and enables rapid arm deformation.
\revise{A compact pump and micro \add{solenoid} valve\add{s}} in the body are connected to their respective inflatable actuators as shown in \figref{all_hardware_new}F. 
Air from the pump as well as air from the bottom airbag can flow into the joint airbag \revise{instantly} due to the \revise{landing impact.}
\add{To evaluate the contribution of the bottom airbag contributes to the deforming time of the robot arm, we compare the time taken to inflate the joint airbags from the pump with and without the pre-inflated bottom airbags.
Detailed experimental procedures are summarized in \appref{bottom faster experiment}. The result indicates that, without the effects of gravity and landing impact, the presence of the bottom airbags accelerates the inflation of the joint airbags by \SI{1.03}{s}.} 

Second, airbags installed at the bottom of the arm can add \revise{bending} torque to the joints of the root segment in \revise{advance}.
\revise{The} airbag expands radially and contracts axially \revise{once the inflating air comes}, so \revise{bending} torque is applied to the joints using the contraction force by fixing the ends of the inflated part of the airbag to the two links, respectively.
This \revise{also leads to the fact that the deformation starts} from the joint at the root of the arm, thereby maximizing the perchable area.
\add{To evaluate this function, we compare the speed of the deforming arm with and without torque applied by fixing the bottom inflatable actuator.
The detailed experimental procedure is summarized in \appref{bottom torque experiment}.
The experiment demonstrated that while increasing bottom actuator pressure slows arm deformation, applying torque mitigates this effect and significantly accelerates deformation, especially at higher pressures. These findings confirm that utilizing torque at the initial stage enhances arm deformation speed efficiency.}

The design of the arm underside airbags requires that they bend without collapsing the flow path to conform to the arm's deformation. 
\revise{Hence,} we develop a one-piece inflatable airbag with an enlarged flow path, as shown in \figref{inflatable}B-2. 
This airbag has partially welded sections that act as joints, allowing bending. The unwelded sections between the welded parts serve as flow paths, while the rest of the airbag serves as expansion sections.
Also, when fabrication, sandwiching non-adhesive PTE tubes during welding enables a larger flow path than welding without sandwiching and ensures the flow path remains unobstructed even when the airbags bend to match the arm morphing as depicted in \figref{inflatable}B-1. \add{Additionally, the expansion section's position is carefully adjusted to minimize interference with the arm’s deforming. Finally, the bottom surface of the bottom inflatable actuator is equipped with high-friction grip material.}

\subsection{\revise{Propeller} Guards for Enhanced Safety}                  
Installing protective components around hazardous areas, such as propellers, not only enhances the safety for individuals but also significantly reduces the psychological burden on humans during the human-drone interaction\cite{abtahi2017drone}.
As shown in \figref{all_hardware_new}G, a protective component designed in the shape of a duct is installed around the propeller to prevent thrust loss. In addition, a hemispherical propeller guard is installed on top of the propeller, which is designed to prevent \revise{airflow resistance} as much as possible.

To evaluate the effect of these propeller guards on thrust, thrust tests are conducted under the following three conditions: with only the motor, with a duct-shaped propeller guard, and a combination of the duct and the hemispherical propeller guard. 
\extraadd{The results, summarized in \tabref{thrust_tests} indicate that the ducted propeller guard generally enhances thrust compared to the motor-only configuration, except for 59.5\%. 
However, when both the ducted and hemispherical propeller guards are combined, the thrust values are slightly reduced compared to the ducted guard alone. 
Despite this, the combined guard configuration still produces higher thrust than the motor-only setup across all PWM values.
This suggests that while the addition of the hemispherical guard leads to some thrust losses, the duct compensates for this as the PWM increases, improving both thrust and safety.}

\extraadd{Furthermore, the required PWM for hovering is calculated, considering the weight of the propeller guards. 
Without guards, the required PWM is 69.1\%, with the ducted guard it is 68. 8\%, and with both guards it is 69. 6\%. The combination of the guards requires slightly higher PWM than the motor-only setup for hovering. 
However, the addition of guards typically increases weight and reduces flight time. For example, when the robot including the weight of both guards is hovered with the motor-only setup, the required PWM increases to 71.1\%. These results suggest that the combined guard configuration can minimize flight time reduction, despite the added weight of the guard.}


Due to the \revise{light-weight design} of the protective components, we use low-density PLA foam material for both the hemispherical propeller guard and the upper part of the duct. This approach results in a 17\% weight reduction per duct compared to \revise{regular PLA material}, and a 33\% weight reduction per \revise{propeller guard, respectively}.

\begin{table}[h!]
\centering
\begin{tabular}{@{}cccc@{}}
\toprule
\textbf{PWM (\%)} & \textbf{Motor Only (N)} & \textbf{Duct-Shaped Guard (N)} & \textbf{Both Guards Combined (N)}    \\ \midrule\midrule
\textbf{59.5}     & 1.03                    & 1.02                      & 1.05                                      \\
\textbf{67}       & 2.74                    & 2.92                      & 2.83                                      \\
\textbf{74.5}     & 4.86                    & 5.23                      & 5.08                                      \\
\textbf{82}       & 6.59                    & 7.14                      & 6.87                                      \\
\textbf{89.5}     & 8.45                    & 9.13                      & 8.99                                      \\ \bottomrule\hline
\end{tabular}
\caption{PWM and Thrust Comparison Under Different Conditions. \extraadd{The thrust values are averages based on four trials, with two CW and two CCW rotations.}}
\label{thrust_tests}
\end{table}

\section{Pneumatic Control System} \label{Pneumatic System}
\subsection{Designing the Pressure for the Inflatable Actuator} \label{determine_p}
\add{The results as stated in \secref{Joint Grasping Force}, indicate that higher joint air pressure increases the grasping force. However, the results of the fatigue test, as detailed in \appref{fatigue_test}, show that the durability of the inflatable mechanism significantly improves when operating at pressures between \SI{40}{kPa} and \SI{50}{kPa}, compared to pressures exceeding \SI{50}{kPa}. Additionally, empirical observations indicate that both inflatable actuators transition from elastic to plastic deformation at approximately \SI{80}{kPa}.
Therefore, to balance durability and grasping performance, we set the maximum inflation pressure to \SI{50}{kPa}, considering a safety factor of \(1.6\) The joint inflatable actuator should be maintained at an optimal pressure between \SI{40}{kPa} and \SI{50}{kPa} during perching.}

\add{On the other hand, the bottom actuator initially functions as a tank, as stated in \secref{bottom_design}. The pre-inflated bottom inflatable actuator equalizes the pressure difference between the joint and the bottom actuators.
Since the stabilized pressure corresponds to the bottom actuator pressure during perching, it is necessary to determine the appropriate bottom initial pressure.
A valve is installed between the bottom actuators and the joint actuators, remaining closed in the initial state, ensuring that only the bottom actuator is pressurized. When the valve opens, air flows from the bottom actuator into the joint actuator.
In this scenario, under isothermal conditions, and based on the principles of mass conservation and Boyle's law, the following relationships are derived for the total mass before and after the valve is opened:
\begin{equation}
\begin{split}
&\frac{(P_{\text{j0}} + P_{\text{atm}}) V_{\text{j0}}}{RT} + \frac{(P_{\text{b0}} + P_{\text{atm}}) V_{\text{b0}}}{RT} \\
=&\frac{(P_{\text{j1}} + P_{\text{atm}}) V_{\text{j1}}}{RT} + \frac{(P_{\text{b1}} + P_{\text{atm}}) V_{\text{b1}}}{RT}
\end{split}\label{isothermal total mass}
\end{equation}
where $P$ represents gauge pressure, $P_{\text{atm}}$ represents atmospheric pressure, $V$ represents volume, $R$ is the gas constant and $T$ is the temperature. 
Also, for the subscripts, $j$ represents the joint inflatable airbag, $b$ represents the bottom inflatable airbag, $0$ represents the closed valve state and $1$ represents the open valve state.
}

\add{From the given conditions, \( P_{\text{j0}} = 0 \), \( P_{\text{b0}} = P_{\text{0}} \) and \( P_1 = P_{\text{j1}} = P_{\text{b1}} \), \equref{isothermal total mass} can be expressed as follows:
\begin{equation}
\begin{split}
    P_1 (V_{\text{j1}} + V_{\text{b1}}) +  P_{\text{atm}} (V_{\text{j1}} + V_{\text{b1}}) \\
    = P_0 V_{\text{b0}} + P_{\text{atm}} (V_{\text{b0}}+V_{\text{j0}})
\end{split}
    \label{P_1}
\end{equation}
}
\add{
Next, we consider the volume of the airbag.
As illustrated in \extraadd{\figref{stable_order_new}B}, when the shape of the airbag is a rectangle with a short side length of $x$ and a long side length of $y$, the airbag after maximum pressurization is approximated as a cylinder with a base circumference of $2x$ and a height of $y$.
The shape of the airbag's base is assumed to be an ellipse during pressurization and becomes a circle at maximum pressurization.
In this case, the volume $V_0$ of the cylinder under max pressure condition is expressed as \(V_0 = \frac{x^2y}{\pi}\). 
}

As the internal pressure decreases, the circular base deforms into an ellipse.
The deformation is characterized by a parameter \( d \), which represents the reduction in the circular base radius.
At maximum pressure \( P=P_{\text{max}} \), the base remains a circle, implying \( d=0 \). At \( P=0\), the base experiences maximum deformation, where \( d=\frac{x}{\pi} \). Accordingly, \( d \) is approximated as linearly dependent on the pressure \( P \), given by the following equation:
\begin{equation}
d = \frac{x}{\pi} \left(1 - \frac{P}{P_{\text{max}}}\right)
\label{d}
\end{equation}
Using this deformation parameter \( d \), the minor axis radius is approximated as \( a =\frac{x}{\pi} - d \), while the major axis radius is approximated as \( b = \frac{x}{\pi} + \frac{\pi - 2}{2} d \) based on the boundary conditions, \( a=0 \) and \( b= \frac{x}{2}\) at \( P=0 \), and \( a=b= \frac{x}{\pi}\) at \( P=P_{\text{max}}\).  
Furthermore, considering that the airbag retains a small residual volume \( V_{\text{res}} \) even at \( 0 \) kPa, the volume \( V(P) \) of the deformed elliptical cylinder is given by the following equation:
\begin{equation}
V(P) = \left(\frac{x}{\pi} + \frac{\pi - 2}{2} d\right) \left(\frac{x}{\pi} - d\right) \pi y + V_{\text{res}} 
\label{Volume}
\end{equation}
Substituting \equref{d} into \equref{Volume}, the equation can be formulated as follows:
\begin{align}
    V(P) &= \frac{x^2 y P}{2 \pi P_{\text{max}}} \left( \pi -\frac{(\pi - 2)P}{P_{max}} \right) + V_{\text{res}}
    \label{V}
\end{align}
To simplify notation, we define the function:
\begin{equation}
F(P) = \frac{P}{2 \pi P_{\max}} \left( \pi - \frac{(\pi - 2) P}{P_{\max}} \right) \label{F_p}
\end{equation}
Considering that the bottom airbag has several inflating parts, the volume of the joint airbags and the bottom airbags can be written as follows by substituting \equref{F_p}, \(x_b\), \(y_b\), \(x_j\), \(y_j\) to \equref{V}:
\begin{align}
    V_{b0} &= \sum_{i=1}^{n_b}\left(\sum_{k} x_{\text{b,k}}^2 y_{\text{b}} F(P_0) + V_{\text{b,res}}\right) \label{V_b0}\\ 
    V_{b1} &= \sum_{i=1}^{n_b}\left(\sum_{k} x_{\text{b,k}}^2 y_{\text{b}} F(P_1) + V_{\text{b,res}}\right) \label{V_b1}\\
    V_{j0} &= \sum_{i=1}^{n_j}{V_{\text{j,res}}} \\
    V_{j1} &= \sum_{i=1}^{n_j}\left(x_j^2 y_j F(P_1) + V_{\text{j,res}}\right) \label{V_j1}
\end{align}

Substituting \equref{V_b0}-\equref{V_j1} into \equref{P_1}, the relationship between \( P_0 \) and \( P_1 \) can be written as followed:
\begin{equation}
\begin{split}
    &\left(P_1+P_{atm}\right)\left(n_j x_j^2 y_j F(P_1) + n_b \sum_{j} x_{\text{b,j}}^2 y_{\text{b}} F(P_1)\right)\\
    &= \left(P_0+P_{\text{atm}}\right) \left( n_b \sum_{j} x_{\text{b,j}}^2 y_{\text{b}} F(P_0)\right) \\
    &\quad+\left(P_0-P_1\right)\left(n_jV_{\text{j,res}} + n_b V_{\text{b,res}}\right)
\end{split} \label{final equ P1}
\end{equation}

By substituting the design parameter, the empirically determined \( P_{\max} \), and the estimated \( V_{\text{res}} \) derived from the measured volume into \equref{final equ P1}, a pneumatic model describing the relationship between \( P_1 \) and \( P_0 \) in the inflatable actuators of the proposed robot is obtained. Details are provided in \appref{pneumatic equation}.
Furthermore, from the approximation line of the measured values shown in \secref{Bottom Grasping Force}, it is observed that when the bottom inflatable actuator pressure reaches \SI{21}{kPa}, the arm exhibits the maximum grasping force. Thus, by setting the target pressure \( P_1 \) to \SI{21}{kPa} after using the bottom inflatable actuators as a tank, and substituting this value into \eqref{final equ P1}, three solutions for \( P_0 \) are obtained: \( P_0 \approx \) \SI{-133.20}{kPa}, \SI{38.20}{kPa} and \SI{213.86}{kPa}.
Among these solutions, the condition \( 0 \leq P_0 \leq P_{\max} = 80 \) uniquely determines the valid value of \( P_0 \) as follows:
\begin{equation}
    P_0\approx 38.20 \label{38}
\end{equation}

\vspace{-28pt}
\subsection{Pneumatic Configuration for Efficient Control for Inflatable Actuators}
\add{The pneumatic device connected to the joint and bottom inflatable actuators consists of a pump, two solenoid valves, and two air pressure sensors. The pump and valves are interconnected as shown in \figref{all_hardware_new}F. 
The operating states of the respective joint and bottom actuators according to the state of the two solenoid valves are detailed in \tabref{solenoid_valve_states_on} and \tabref{solenoid_valve_states_off}.
There are two major features of pneumatic devices configuration.}

\add{The first feature is that the air outlet of the bottom inflatable actuator is connected to the air inlet of the joint inflatable actuator thought of the solenoid valve. 
When the solenoid valve is turned on, the pressure of the joint airbags is exhausted. 
Furthermore, a check valve attached to the outlet of the bottom actuator allows air to flow from the bottom actuator to the joint actuator when the bottom pressure is higher.
Conversely, when the joint actuator has higher pressure, the check valve prevents backflow, enabling it to maintain that pressure, as shown in \tabref{solenoid_valve_states_on}. 
This configuration allows the bottom airbag to function as a reservoir, allowing rapid airflow into the joint actuator.} 

The second feature is that actuator pressure is maintained even after the pump and valves are turned off, as shown in \tabref{solenoid_valve_states_off}, indicating that no additional pressure is required and energy consumption is minimized during perching and grasping the human arm, as long as the actuators remain intact.

\begin{table}[h!]
\centering
\begin{tabular}{ll|ll|ll}
\cline{3-4}
&      & \multicolumn{2}{c|}{\textbf{SV2}} &  &  \\ \cline{3-4}
&      & \multicolumn{1}{c|}{\textbf{ON}}  & \multicolumn{1}{c|}{\textbf{OFF}}                                                                       &  &  \\ \cline{1-4}
\multicolumn{1}{|c|}{\multirow{2}{*}{\textbf{SV1}}} & \multicolumn{1}{c|}{\textbf{ON}}  & \multicolumn{1}{l|}
{\begin{tabular}[c]{@{}l@{}}\textbf{J}: Exhaust \\ \textbf{B}: Intake(Pump) \end{tabular}} 
& \begin{tabular}[c]{@{}l@{}}
\textbf{J > B} \\ 
\hspace{3.0mm}J: Hold \\ 
\hspace{3.0mm}B: Intake(Pump) \\ 
\textbf{J <= B} \\ 
\hspace{3.0mm}J: Intake(Pump) \\ 
\hspace{3.0mm}B: Intake(Pump)
\end{tabular} &  &  \\ \cline{2-4}
\multicolumn{1}{|c|}{}                     
& \multicolumn{1}{c|}{\textbf{OFF}} 
& \multicolumn{1}{l|}{\begin{tabular}[c]{@{}l@{}}
\textbf{J}: Intake + Exhaust \\ 
\textbf{B}: Hold
\end{tabular}} & \begin{tabular}[c]{@{}l@{}}
\textbf{J > B} \\ 
\hspace{3.0mm}J: Intake(Pump) \\ 
\hspace{3.0mm}B: Hold \\ 
\textbf{J <= B} \\ 
\hspace{3.0mm}J: Intake(Pump + Bottom) \\ 
\hspace{3.0mm}B: Exhaust(Joint)
\end{tabular} &  &  \\ \cline{1-4}
\end{tabular}
\caption{Actuator States Based on Solenoid Valve Status When the Pump Work On \\
\textbf{J} represents the Joint inflatable actuator, while \textbf{B} represents the Bottom inflatable actuator.}
    \label{solenoid_valve_states_on}
\end{table}

\begin{table}[h!]
\centering
\begin{tabular}{ll|lc|ll}
\cline{3-4}
&                          & \multicolumn{2}{c|}{\textbf{SV2}}   &  &  \\ \cline{3-4}
&                          & \multicolumn{1}{c}{\hspace{1.0cm}\textbf{ON}}    & \textbf{OFF}   &  &  \\ \cline{1-4}
\multicolumn{1}{|c|}{\multirow{2}{*}{\textbf{SV1}}} 
& \multicolumn{1}{c|}{\textbf{ON}}  
& \multicolumn{2}{l|}{\begin{tabular}[c]{@{}l@{}}
\textbf{J}: Exhaust \\ 
\textbf{B}: Hold 
\end{tabular}} &  &  \\ \cline{2-4}
\multicolumn{1}{|c|}{} 
& \multicolumn{1}{c|}{\textbf{OFF}} 
& \multicolumn{2}{l|}{\begin{tabular}[c]{@{}l@{}}
\textbf{J > B} \\ 
\hspace{3.0mm}\textbf{J}: Hold \\ 
\hspace{3.0mm}\textbf{B}: Hold (Different Pressure) \\ \textbf{J <= B} \\ 
\hspace{3.0mm}\textbf{J}: Hold \\ 
\hspace{3.0mm}\textbf{B}: Hold (Same Pressure) \end{tabular}}&  &  \\ \cline{1-4}
\end{tabular}
\caption{Actuator States Based on Solenoid Valve Status When the Pump Work Off \\
\textbf{J} represents the Joint inflatable actuator, while \textbf{B} represents the Bottom inflatable actuator.}
    \label{solenoid_valve_states_off}
\end{table}

\subsection{Pneumatic Control for Adjusting Pressure}
\add{To ensure precise and stable pressure control, a proportional feedback control mechanism is implemented. Each pressure sensor measures the internal air pressure of each actuator and provides feedback to the pump output. 
Specifically, the PWM signal sent to the pump ($PWM_\text{output}$) is expressed as follow:
\begin{equation}
    PWM_{\text{output}} = (P_{\text{target}} - P_{\text{cur}}) \times K_p + PWM_{\text{min}}
\end{equation}
where \(P_{\text{target}}\) represents the desired air pressure, \(P_{\text{cur}}\) is the current air pressure, \(K_p\) is the proportional gain, and \(PWM_{\text{min}}\) is the minimum PWM signal required to activate the pump.}

\add{This control strategy not only enables precise air pressure adjustment but also acts as a fail-safe mechanism, protecting the system in two key ways. 
First, it prevents over-pressurization that could damage the actuators, ensuring system stability and longevity. 
Second, it allows the system to compensate for minor actuator damage, such as small punctures or leaks, maintaining functionality even under partial failure conditions.}

\add{Furthermore, according to the states detailed in \appref{planning}, \tabref{switching_airflow} illustrates the pump and valves switching, along with the state transitions of the inflatable actuators.
In the robot's task planning, all state transitions, except for the switch from Approach to Reach, are determined by pressure conditions. To ensure safety, if the system fails to maintain sufficient pressure due to air leakage, the next transition will not be triggered.}

\begin{table}[h!]
  \centering
  \begin{tabular}{c|c|c|c|l}
    \toprule
    & \textbf{Pump} &\textbf{Valve1} &\textbf{Valve2} &\textbf{Airbag State}\\ \hline\hline
    \textbf{Approach} & ON(Needed)  & ON  &ON  & Bottom actuator pre-inflated\\ \hline
    \multirow{2}{*}{\textbf{Reach}}  
    & ON  & ON   & ON & Bottom actuator is inflated at \(P_0\)\\ \cline{2-5}
    & OFF & OFF  & OFF & Joint actuator is inflated until maintaining arm rigidity \\ \hline
    \multirow{2}{*}{\textbf{Perch}}
    & ON(Maximum) & OFF & OFF & Joint actuator is fully pressurized at \SI{50}{kPa}\\ \cline{2-5}
    & OFF & OFF  & OFF& Pressure hold \\ \hline
    \textbf{Deperch} & OFF & OFF & ON & Air Exhausted  \\ \bottomrule
  \end{tabular}
  \caption{Solenoid Valves and Pump in Each State}
  \label{switching_airflow}
\end{table}


\section{\revise{Modeling and Control}}\label{control}
\renewcommand{\W}{\{W\}}
\newcommand{\G}{\{G\}}
\newcommand{\R}{\bm{R}_{\G}}
\newcommand{\E}{\bm{E}}
\newcommand{\tr}{\mathrm{T}}
\newcommand{\force}{\bm{f}}
\newcommand{\torque}{\bm{\tau}}
\newcommand{\thrust}{\bm{\lambda}}
\newcommand{\pos}{\bm{r}}
\newcommand{\ui}{\bm{u}_i}
\newcommand{\p}{\bm{p}_i}

\newcommand{\wphi}{\phi}
\newcommand{\wtheta}{\theta}
\newcommand{\wpsi}{\psi}
\newcommand{\dotphi}{\dot{\phi}}
\newcommand{\dottheta}{\dot{\theta}}
\newcommand{\dotpsi}{\dot{\psi}}
\vspace{-35pt}
\extraadd{
\subsection{Thrust Condition for Stable Hovering}\label{design_req}
}
Modeling the kinematics and dynamics of the flexible arms is crucial for understanding the motion of the entire aerial robot.
Ryll et al.\cite{ryll2022smors} employ Piecewise Constant Curvature (PCC) approximation and utilize extended state representations to approximate completely flexible arms as a combination of translational and rotational joints.
However, such a nonlinear modeling would make the control framework complex. 
Therefore, we focus on the dominant situation of the proposed unilateral flexible arm while sufficient thrust is provided. 
We assume that the unilateral flexible arm can be considered a rigid model during flight with sufficient as shown in \extraadd{\figref{stable_order_new}C}. 
Let $m$, $l$, $M_{\text{A}}$, $R_{\text{A}}$, $\lambda_{\text{hover}}$ and $\rho$ denote the total mass of the robot, length from root to the rotor mount, bending moment at the base of the arm, force at the arm base, thrust of each rotor during hovering, and mass per unit arm length, respectively. Note that $l$ is not identical to the arm length because the rotor is mounted at the center.
Then, the total mass $m$ and hovering thrust $\lambda_{\text{hover}}$ can be written as follows:
\begin{align}
  m &= m_{\text{body}} + 4 m_{\text{arm}} + 4 m_{\text{rotor}}, \label{eq:weight_of_the_machine}\\
  4 \lambda_{\text{hover}} &= mg, \label{eq:arm_lambda_max}
\end{align}                  
where $m_{\text{body}}$, $m_{\text{arm}}$ and $m_{\text{rotor}}$ represent the body weight, arm weight, and rotor weight, respectively.
Besides, the moment equilibrium equation around the arm base can be written as follow:
\begin{equation}
  - M_{A} + (\lambda_{\text{hover}} - m_{\text{rotor}} g ) l  - \int^{l}_{0} \rho g x dx = 0.  \label{eq:arm_moment}\\
\end{equation}

Considering the following characteristics of the arm $M_{\text{A}} \geq 0\label{eq:arm_moment_A}$, the hovering thrust $\lambda_{hover}$ can be given by following:
\begin{align}
  \lambda_{\text{hover}} & \geq \frac{m_{\text{arm}}g}{2} + m_{\text{rotor}} g, \label{eq:arm_thrust_geq}
\end{align} 
where $\rho  l\approx m_{\text{arm}}$.

In addition, by substituting \equref{eq:arm_thrust_geq} into \equref{eq:weight_of_the_machine} and \equref{eq:arm_lambda_max}, the relationship of mass components can be written as follow:
\begin{align}
  m_{\text{body}} + 2 m_{\text{arm}}  \geq 0.
    \label{eq:arm_moment_geq}
\end{align}
Given that $m_{\text{body}}$ and $m_{\text{arm}}$ are positive, the above equation is always established, indicating that the robot can always keep stretched without bending in the case around the hovering state.

\vspace{-25pt}
\extraadd{
\subsection{Standard Quadrotor-Based Model and Control}\label{robustness}
As discussed in \secref{design_req}, when sufficient thrust is applied, the unilateral flexible arm behaves similarly to a rigid structure due to its structural bistability. 
This allows the unilateral flexible arm to be approximated as a rigid body near the hovering condition, making it possible to apply standard quadrotor dynamics modeling and flight control.}
\revise{In contrast, if the near-hovering conditions are not satisfied, the arm may unexpectedly bend downwards because of the insufficient thrust force, resulting in unstable motion in midair.
For example, when the robot is required to generate a relatively large moment around the yaw axis, a pair of rotors diagonally aligned will decrease the thrust and thus the drag torque as shown in \figref{stable_order_new}D. Then the corresponding arms will bend downwards, which directly leads to descent because of the decrease in the total force against gravity.}


If the deforming or deformation of the flexible structure can be quantitatively measured or estimated, the whole dynamics model can be observed even though it is highly nonlinear. Then, the model-based control methodology, such as Model Predictive Control, can be applied to handle the nonlinearity. 
However, the computational costs not only for control but also for state estimation can be significantly high for on-board processor with limited computational resource. 
Hence, it is desired to adopt a simple control method based on effective approximation. 

Focusing on the typical deviation moment as depicted in \textcircled{2} of \figref{stable_order_new}\add{D}, the deforming arms change the direction of the corresponding thrust forces. Except for the force along the z-axis, the total force or moment regarding other axes, especially the roll and pitch axes, has a similar behavior with the hovering situation if we follow the nominal allocation-based control method for quadrotor as presented by T. Lee et al.\cite{CDC2010-SE3}. 
This further indicates the recovery ability to stretch back the arm to the rigid mode as depicted in \textcircled{3} of \figref{stable_order_new}D, even if the control gains do not change from the phase of \textcircled{1}. 
Eventually, the robot is expected to converge to the stable hovering situation as shown in \textcircled{4}.

\extraadd{
Therefore, in this work, we introduce a standard quadrotor modeling to describe the dynamics in the near-hovering state, as presented in \appref{flight_control}. 
It is notable that the allocation matrix should vary according to the arm deformation. 
However, we approximate it as constant by leveraging the structural bistability of the arms, which maintains a rigid-like configuration under sufficient thrust.
This approximation simplifies the dynamics model, making it much more simple to handle for subsequent control method.
Then, a unified control strategy can be applied across all flight conditions.
Specifically, we employ a dual-loop LQI-based controller \cite{hydrus_mbzirc}, as detailed in \appref{attitude_control}.
}  
\section{Results} \label{experiments}

\subsection{Overall Configuration}
\begin{figure}[h]
    \centering
    \includegraphics[width=1.0\columnwidth]{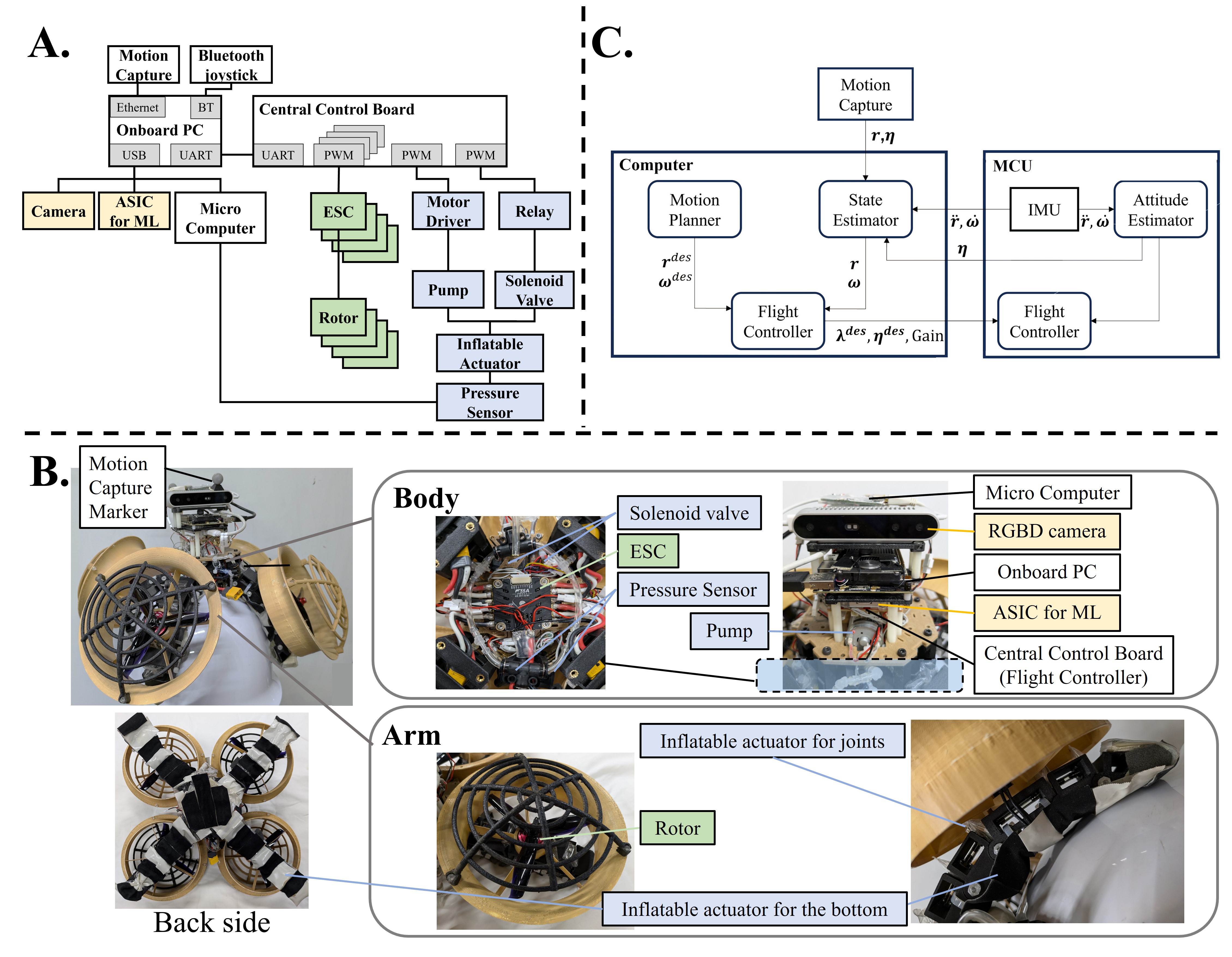}
    \caption{Hardware diagram of the prototype of deformable aerial robot.
    (A) diagram of the hardware system.
    (B) overall flight system diagram of the control system.
    (C) hardware configuration.
    }
    \label{hardware_system}
\end{figure}

The overall hardware system configuration is illustrated in \figref{hardware_system}A.
The total weight of the aerial robot in this study is \SI{1.444}{kg}, with dimensions of $\SI{350}{mm} \times \SI{350}{mm} \times \SI{250}{mm}$. 
The details of the weight are shown in \tabref{components_weights}.
The aerial robot consists of two systems: the flight communication system and the grasping pneumatic system.
Regarding the flight system, rotors and propellers, with 5-inch propellers, are mounted at the center of the arms.
The overall flight control system is designed, as depicted in \figref{hardware_system}B. 
Additionally, ESC \extraadd{(T-MOTOR F55A PROII 6S 4-IN-1 ESC)}, an onboard PC (\revise{Khadas VIM4 Amlogic A311D}) and a central control board (STM32H7) are mounted in the body, as shown in \figref{hardware_system}C.
In the pneumatic system, the pump \extraadd{(TKC27-1-3-0001)} is controlled via PWM through the central control board and motor drivers, and the solenoid valves \extraadd{(SMC S070C-6CC-32)} are controlled via PWM through relay terminals from the central control board.
The pneumatic pressure sensors \extraadd{(MPX5700DP)} connect to the microcomputer \extraadd{(Arduino\textsuperscript{\textregistered}Nano)}, transmitting analog signals.
Furthermore, we characterize this study robot that can perch on a human arm in comparison to a variety of other perching robots as detailed in \appref{comparison}.

\begin{table}[h!]
    \centering
    \begin{tabular}{@{}l|l|r@{}}
        \toprule
        \textbf{Category} & \textbf{Component} & \textbf{Weight (kg)} \\ \midrule\midrule
        \multirow{14}{*}{Body}
            & Flame & 0.145 \\ 
            & Onboard PC (Khadas VIM4 Amlogic A311D)& 0.059 \\
            & Central Control Board (STM32H7)& 0.016 \\
            & Microcomputer (Arduino\textsuperscript{\textregistered} Nano)& 0.008 \\
            & RGBD Camera (Intel\textsuperscript{\textregistered} RealSense\texttrademark{} Depth Camera D415)& 0.063 \\
            & Edge AI Device (Coral Edge TPU)& 0.020\\
            & ESC (T-MOTOR F55A PROII 6S 4-IN-1 ESC
)& 0.016 \\
            & \underline{Pump (TKC27-1-3-0001)}& 0.061 \\
            & \underline{Air Pressure Sensor (MPX5700DP)}& 0.006 *2 \\
            & \underline{Solenoid Valve (SMC S070C-6CC-32)}& 0.008 *2 \\
            & \underline{Check Valve (PISCO CVLU4-4)} & 0.006 \\
            & \underline{Bottom Inflatable Actuator} & 0.011 \\
            & Others & 0.103 \\ 
            & Battery (4S 14.8V) & 0.134 *2 \\ \hline
        & & \textbf{0.536 + 0.268} \\ \midrule
        \multirow{7}{*}{Arm} 
            & Motor (ARRIS S2205 2300KV) and Propeller & 0.036 *4 \\
            & Ducted Propeller Guard & 0.031 *4 \\
            & Hemisphere Propeller Guard & 0.014 *4 \\
            & Unilateral Flexible Arm & 0.036 *4 \\
            & \underline{Joint Inflatable Actuator} & 0.002*5 *4 \\
            & \underline{Bottom Inflatable Actuator} & 0.009 *4 \\ 
            & Others & 0.024 *4 \\\hline
            & & \textbf{0.160 *4} \\
            \midrule
        \textbf{Total} &  & \textbf{1.444} \\ \bottomrule
    \end{tabular}
    \caption{Components and Weights of the Robot, \extraadd{The underlined components are required for the morphing, accounting for 12.6\% (\SI{0.182}{kg}) of the total weight.}}
    \label{components_weights}  
\end{table}

\vspace{-25pt}
\add{\subsection{Grasping Experiment}}
\subsubsection{Effect of Joint Inflatable Actuator on Grasping Force}\label{Joint Grasping Force}
\add{We conducted grasping force experiments using various objects. 
The results are shown in \figref{evaluation_grasping_force}A, with the experimental setup and a sample force measurement presented in \figref{evaluation_grasping_force}B-1 and \figref{evaluation_grasping_force}B-2. Details are summarized in \appref{grasping}.}
\add{To evaluate the effect of joint inflatable actuator on grasping force, we set the bottom actuator pressure to \SI{0}{kPa} and varied the joint actuator pressure during grasping a human arm model. The results demonstrate that increasing the pressure in the joint inflatable actuator leads to a higher grasping force, as shown in \figref{evaluation_grasping_force}C.}

\subsubsection{Effect of Bottom Inflatable Actuator on Grasping Force} \label{Bottom Grasping Force}
\add{We conducted experiments using an arm model and a box with a height of \SI{50}{mm}, keeping the joint inflatable actuator pressure at \SI{50}{} ± \SI{4}{kPa} while varying the bottom actuator pressure from \SI{0}{} to \SI{50}{kPa}, as shown in \figref{evaluation_grasping_force}D.
The results show that grasping force improves when using the bottom actuator, regardless of its pressure.
The highest grasping force was achieved when grasping a H\SI{50}{mm} box, reaching \SI{45.17}{N} at a joint pressure of \SI{47}{kPa} and a bottom pressure of \SI{0}{kPa}. When grasping a human arm model, the maximum grasping force was \SI{41.65}{N} at a joint pressure of \SI{50}{kPa} and a bottom pressure of \SI{19}{kPa}.}

\add{For the  H\SI{50}{mm} box, grasping force was highest when the bottom pressure was between \SI{0}{}–\SI{10}{kPa}, whereas for the human arm, it peaked around \SI{20}{kPa}. In both cases, grasping force is greater at lower bottom inflatable actuator pressures compared to higher pressures. 
This is likely because increased pressure makes the bottom actuator stiffer, reducing its ability to conform to the object's surface. Additionally, as shown in \figref{evaluation_grasping_force}E, the contact points between the end effector and the object shift upward, which may alter the direction of the applied force and further decrease grasping force. Further investigation is needed to verify this effect.}

\add{Furthermore, the difference in the optimal bottom actuator pressure for achieving maximum grasping force is likely due to the shape adaptability of the unilateral flexible arms. 
When grasping a H\SI{50}{mm} box, the arms already conform well to the object, so further inflation of the bottom actuator reduces conformity, decreasing grasping performance. In contrast, when grasping a human arm model, the bottom actuator helps fill the gap between the arms and the object, increasing the contact area and enhancing the grasping force.}

\subsubsection{Effect of Object Shape and Size on Grasping Force with Identical Actuator Pressures}
\begin{figure}[h]
    \centering
    \includegraphics[width=1.0\columnwidth]{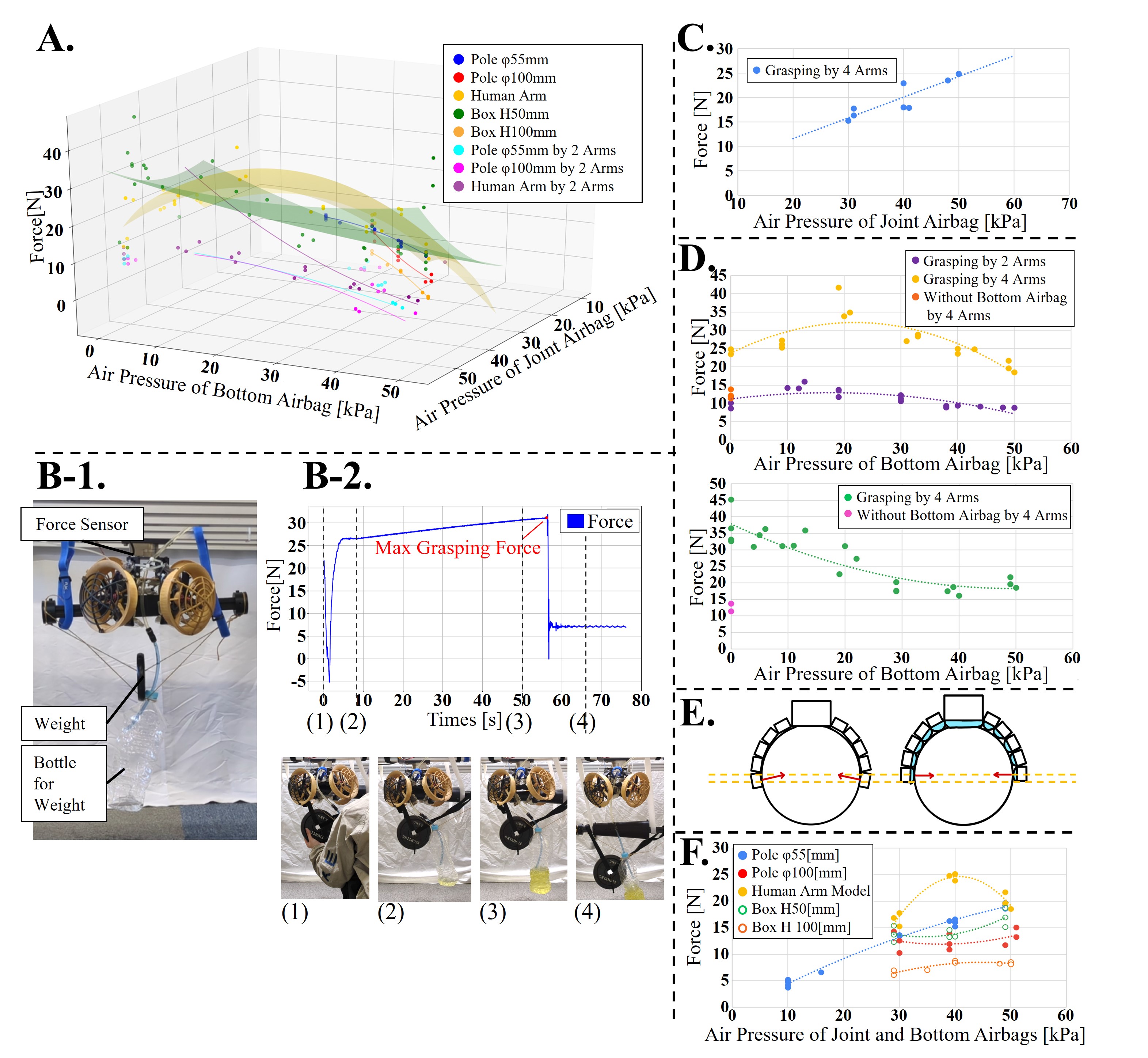}
    \caption{\add{(A) Plot of grasping force results from all experiments.
(B-1) Experimental setup for grasping force measurement.
(B-2) Time-series plot of grasping force during the experiment. The maximum grasping force is determined as the force just before reaching the peak, right before the object starts to slip.
(C) Plot of maximum grasping force when grasping the arm model, with the bottom inflatable actuator pressure set to \SI{0}{kPa} and the joint inflatable actuator pressure varied.
(D) Plot of maximum grasping force when the joint inflatable actuator pressure is fixed at \SI{50}{} ± \SI{4}{kPa}, while the bottom inflatable actuator pressure is varied. The upper plot represents the results for the arm model, while the lower plot corresponds to the H\SI{50}{mm} box.
(E)
(F) Plot of grasping force when various objects were grasped with identical pressures in both the joint and bottom inflatable actuators.}}
    \label{evaluation_grasping_force}
\end{figure}

\add{We evaluated grasping force with identical pressures in the joint and bottom actuators while grasping various objects, as shown in \figref{evaluation_grasping_force}F.}
\add{The highest grasping force (\SI{25.10}{N}) was observed when grasping the arm model at \SI{40}{kPa} for both actuators, with performance decreasing at \SI{50}{kPa}. In contrast, cylindrical and rectangular objects reached maximum grasping force at \SI{50}{kPa}.}

\add{The results indicates that the variation in optimal pressure depending on the object suggests that the degree to which the bottom actuator hinders grasping differs based on the object’s shapes and properties.
For example, in the case of the H\SI{50}{mm} box, grasping force remains relatively unchanged within a bottom actuator pressure range of \SI{30}{kPa}–\SI{50}{kPa}. However, for the human arm model, grasping force decreases as the bottom actuator inflates within the same pressure range. This suggests that for box-like objects, the joint inflatable actuator contributes more to grasping force, leading to the highest force. In contrast, for the arm model, the grasping force peaks where the effects of the joint and bottom actuators are balanced.}
\add{Additionally, as object size increased (e.g.,  \SI{100}{mm} rectangular object), the arms could not fully wrap around it, leading to a decrease in grasping force.}

\subsubsection{Grasping \add{Stability} Experiment on Various Objects}
\add{To verify the grasping capability of the unilateral flexible arm in practical scenarios,} we conducted experiments \add{to assess whether the robot could maintain a stable grasp on various objects}, such as a helmet, a basketball, and a chair, when the held object experienced movement, as shown in \figref{bunmawashi}A.
Subsequently, the held object was tilted in the \extraadd{$y$} direction until the robot fell to determine the maximum angle at which gripping could be maintained.
\revise{There were differences in the maximum angle at which gripping could be sustained.} For example, in the case of the helmet, the maximum \extraadd{pitch} angle was approximately \extraadd{\SI{1.12}{rad}}, while perching on a cylindrical object, gripping could be sustained up to approximately \extraadd{\SI{1.35}{rad}}.

The ability to grasp various objects and hold on the tilted objects proved \add{the practical effectiveness} of both the joint inflatable actuator that deforms the arm to fit the object's shape and the bottom inflatable actuator.

\subsubsection{Grasping Experiment on Swinging Human Arm}
\begin{figure}[h!]
    \centering
    \includegraphics[width=1.0\columnwidth]{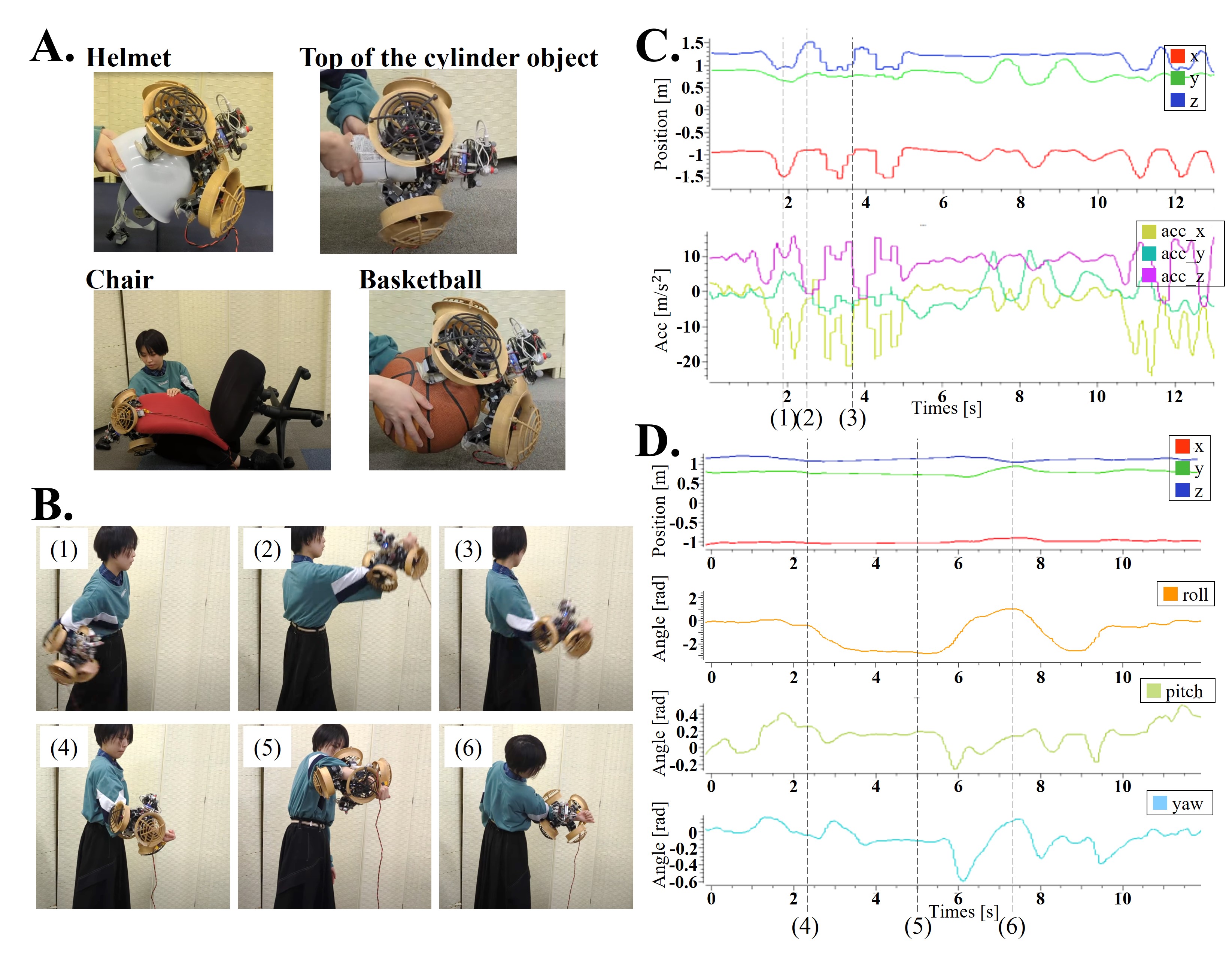}
    \caption{Experiments in clinging to \revise{various} objects and people
    (A) clinging to the various objects that tilts in steps after grasping them.
    (B) clinging to the arm of a person who is swinging, supinating and pronating.
    (C) plots of position and acceleration data while grasping a human arm swinging in the $x$ direction.
    (D) plots of position and angle data while clinging to a human arm spinning in the roll direction.}
    \label{bunmawashi}
\end{figure}

As depicted in \figref{bunmawashi}B, we conducted experiments to assess the gripping capability of this robot on human \revise{body}.
With its fully expanded joints and bottom inflatable actuators, it clung to a human arm while swinging along the $x$ axis and internal/external rotation in the roll direction.
As shown in \figref{bunmawashi}C, when the acceleration along the $x$ axis remained within the range of \extraadd{\SI{0.5}{m/s^2}} to \extraadd{\SI{-14.7}{m/s^2}}, it could cling without being shaken off.
Additionally, according to the experimental results depicted in \figref{bunmawashi}D, it was found that even with a maximum roll of approximately \extraadd{\SI{1.0}{rad}} and a minimum of approximately \SI{-2.8}{rad}, the robot could continue to grip onto the human arm.
While clinging to the human arm, even though \add{the solenoid valves' power and} the pump's power is off, the air pressure inside the inflatable actuators can be maintained constant, \revise{which indicating the advantage of energy saving.}

\subsection{Evaluation of Flight Robustness}
\begin{figure}[h!]
    \centering
    \includegraphics[width=0.7\columnwidth]{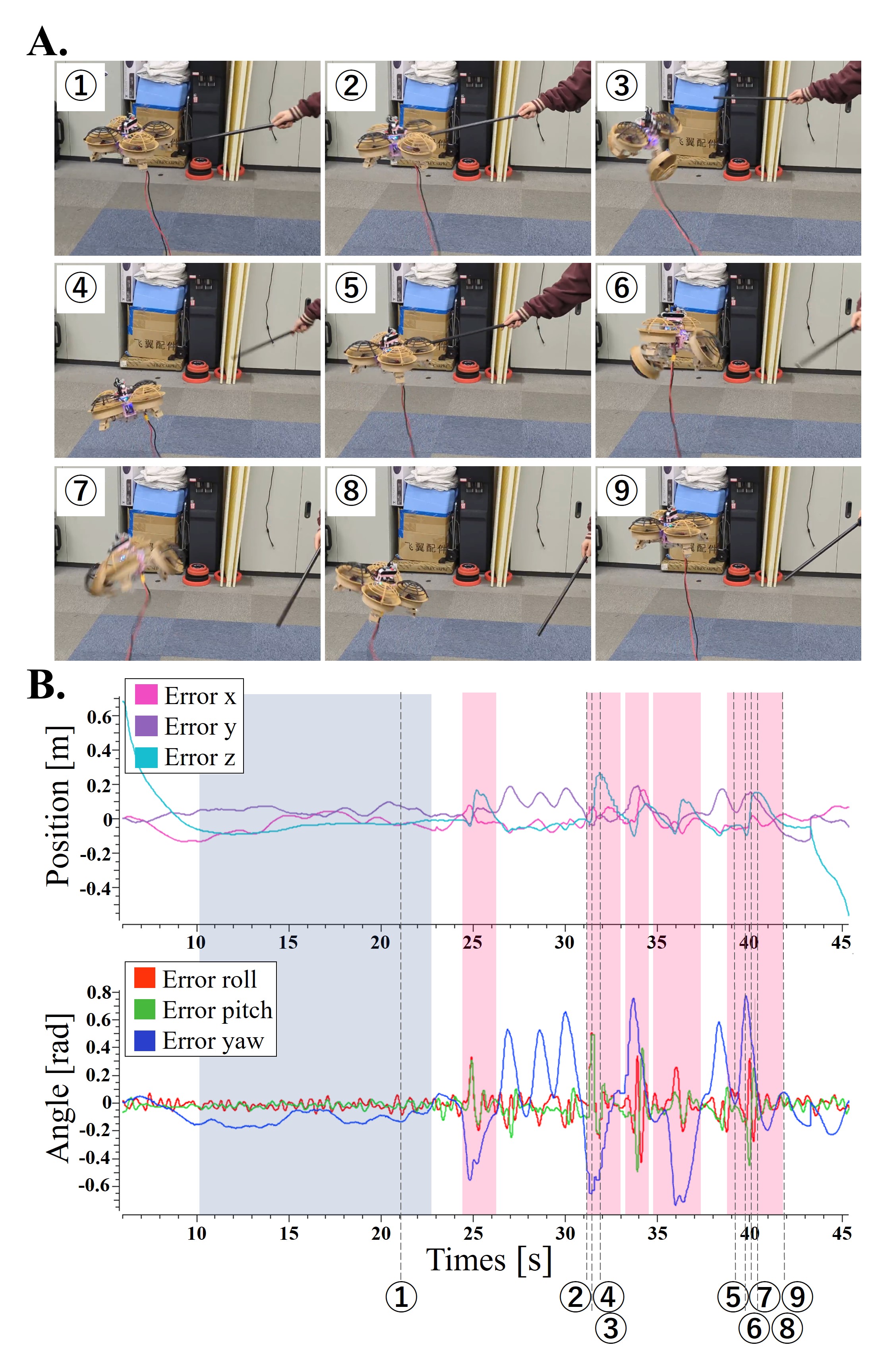}
    \caption{Flight stability \revise{evaluation}.
    (A) snapshots of the experiment of attitude recovery when the arms flex in flight.
    (B) plots of the experiment. Above: errors of Position in the $x$, $y$ and $z$ axes. Below: errors of roll, pitch and yaw angles. The blue region in the figure is the duration of hovering and the red region is the duration of adding yaw angle disturbance.}
    \label{Robust}
\end{figure}

We conducted experiments by pushing the robot while hovering with a stick to introduce disturbances in the yaw direction and evaluate its robustness.
As shown in \figref{Robust}A, when the disturbance causes the rotor's thrust to fall below the minimum required to maintain the arm's rigidity, the arm hung down, leading to the deviation in both altitude and attitude motion.
While the RMSE during hovering was [0.019 0.016 0.013]\si{m} and [0.031 0.017 0.050]\si{rad} as depicted in \figref{Robust}B, after imposing disturbances, the roll and pitch errors significantly increased within the range of \revise{\SI{0.4}{rad}} to \SI{-0.5}{rad}.
However, although the robot \revise{descended} in the negative \revise{$z$}-direction, \revise{it could rapidly converge to the target height again owing to the proposed flight control.
Furthermore, it was confirmed that even if the aerial robot's posture became unstable, the rotational motion along the roll and pitch directions did not diverge. This indicates that the thrust generated by the rotors on the deforming arms also contributed to attitude control.}
\revise{As a result, this robot demonstrated a desired robustness against the disturbance despite having flexible arms}, thereby validating the hypothesis and the effectiveness of the control methods described in \secref{control}.

\subsection{In-flight Perching on Objects and Human}
\subsubsection{\revise{Perching on Objects}}
To verify perching performance in the environment, we conducted a series of experiments involving flight, perching and deperching on a cylindrical structure.
The perch target had a diameter of \SI{150}{mm} and was constructed by wrapping an aluminum frame with cushioning material.
The aerial robot in this experiment was equipped only with the joint inflatable actuator, with sponges installed on the bottom for shock absorption.

\begin{figure}[h!]
    \centering
    \includegraphics[width=1.0\columnwidth]{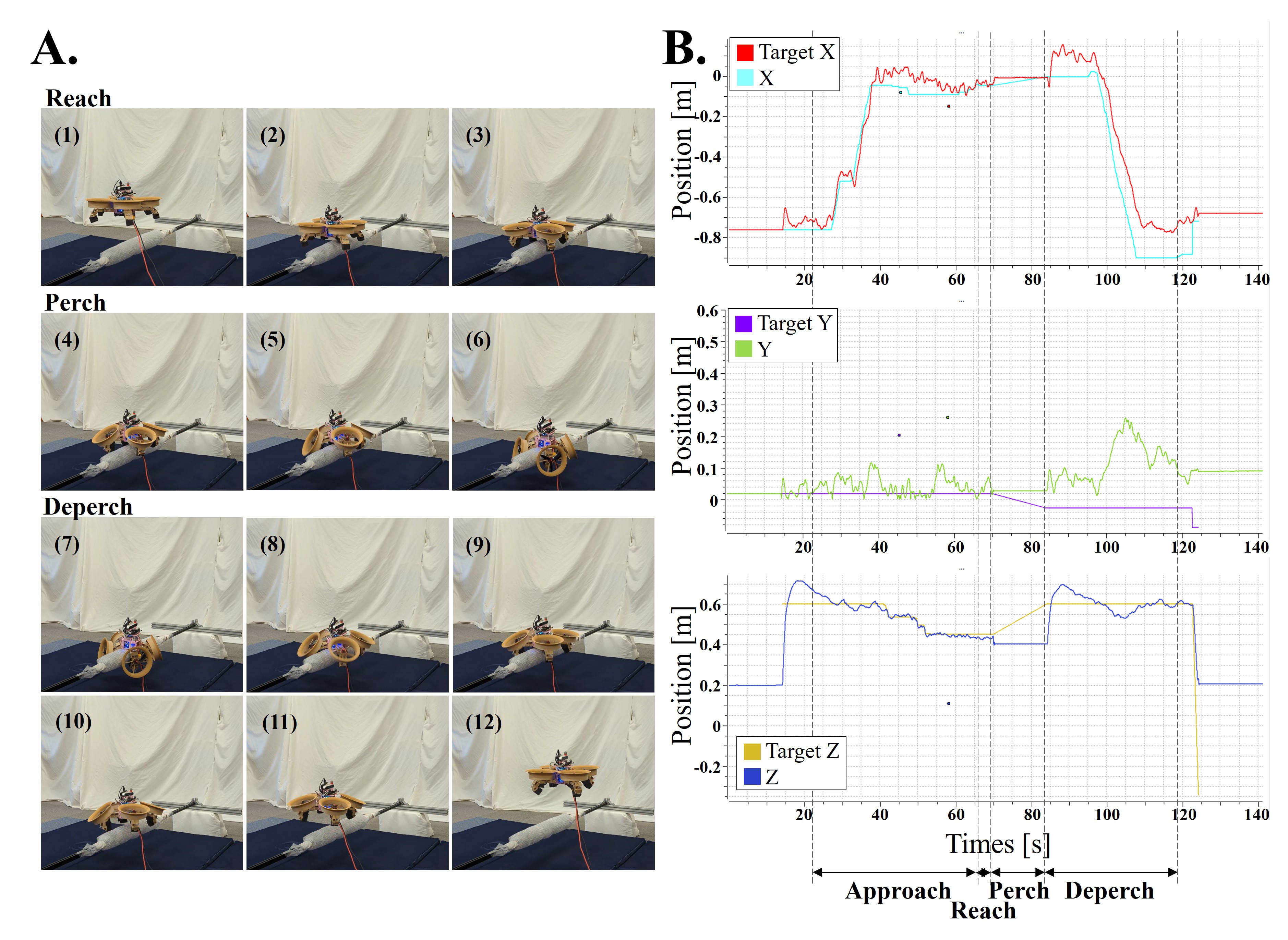}
    \caption{Experiment of perching on cylinder beam.
    (A) snapshots of the experiment.
    (B) plot of the target position and the measured values.}
    \label{object_perching}
\end{figure}

As shown in \figref{object_perching}, the robot \revise{succeeded} to perch and deperch automatically.
\revise{It took approximately \SI{3.6}{s} from inflation to perching, and \SI{7.0}{s} from  propeller activation to takeoff. 
The robot dropped and perched from approximately \SI{0.05}{m} from the object.
In deperch state, the robot moved up to \SI{0.1}{m} in the z-direction from the target position, but then stabilized its position.}
Additionally, the measured position closely followed the desired value, demonstrating the successful execution of the task.

\subsubsection{\revise{Perching on Human Arm}}
\begin{figure}[h!]
    \centering
    \includegraphics[width=1.0\columnwidth]{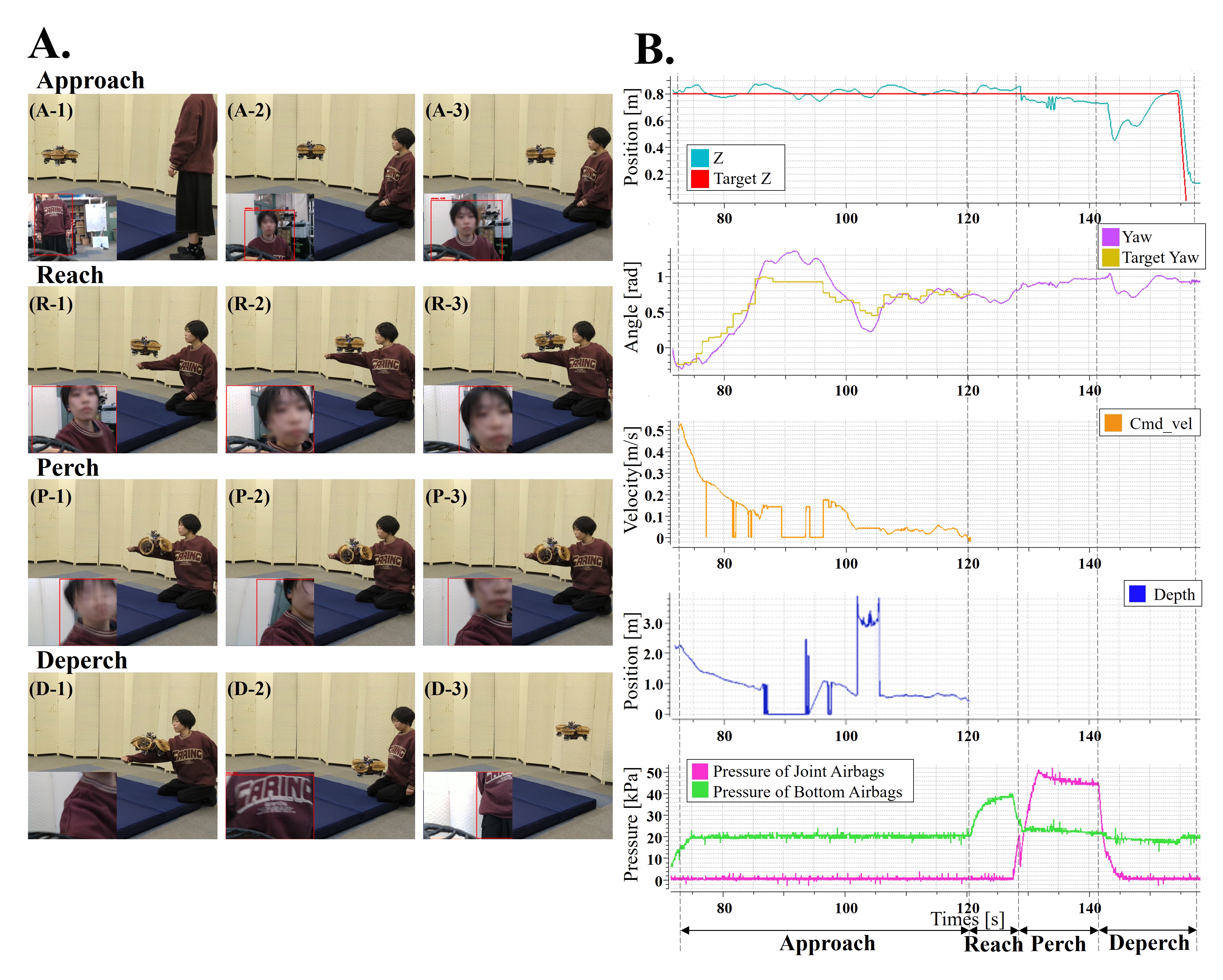}
    \caption{Experiment of perching on a human arm from in-flight situation.
    (A) Snapshots of the experiment of perching on a human arm. The small sub-image below the left of each image is the human recognition result using an onboard RGB camera image.
    (B) plots of the target and measured \add{z position and yaw angle, along with the desired velocity and estimated distance towards the target person and the pressure of joint and bottom inflatable actuators}}
    \label{human_perching}
\end{figure}

To evaluate the performance of perching-based human-aerial-robot interaction, the robot demonstrated \revise{a sequential action}: approaching the human, perching on a human arm, and taking off from a human arm \revise{as shown in \figref{human_perching}}.

The perching motion was achieved using the Finite State Machine (FSM) as detailed in \appref{planning}, where state transitions occur autonomously based on human recognition data and actuator pressure feedback.
The initial takeoff \add{was} commanded manually, while all other actions were performed autonomously.
In addition, each state and the pressures of each inflatable actuator are summarized in \tabref{pressure_table}.

Firstly, in \add{\textbf{Approach State}, the robot could track and approach the moving human using velocity-based PD control, with the yaw angle actively adjusted to maintain orientation toward the human, }as illustrated in \figref{human_perching}A (A-1) to (A-3).
As shown in the command velocity and depth graph in \figref{human_perching}B, the estimated distance decreases, \revise{which gradually reduced} velocity. 
\add{Additionally, during this phase, the bottom inflatable actuator was maintained at \SI{20}{kPa} to absorb impact in case of an unexpected fall.}

Then, when the robot reached a certain distance from the target person (i.e., \SI{0.5}{m}), it hovered in place in \add{next state}.
Even when only a part of the human was visible, it could still recognize the person and acquire the estimated distance, as depicted in \figref{human_perching}A(R-3).

\add{In \textbf{Reach State}, the robot hovers in preparation for perching, while a human actively extend their arm, similar to a falconer inviting a bird to perch, as shown in \figref{human_perching}A(R-2).}
Prior to this, the bottom inflatable actuator is pre-pressurized to \SI{40}{kPa} to enable rapid inflation of the joint actuator during perching. This value is based on the optimal pressure for human perching derived in \equref{38}.
\add{Once the bottom actuator reaches \SI{40}{kPa}, the solenoid valve opens, allowing air to flow into the joint actuator until it reaches \SI{19}{kPa}, the maximum pressure needed to maintain the arm rigidity.
At this point, the propellers stop, initiating a free fall and arm deformation.
If the joint actuator does not reach \SI{19}{kPa}, the pump automatically activates \SI{1.5}{s} after the valve opens to compensate.}

\add{According to the result shown in \figref{human_perching}B, \SI{1.4}{s} after the bottom actuator reached \SI{40}{kPa}, the joint actuators reached \SI{19}{kPa}, while the bottom actuator was at \SI{27}{kPa}. At this point, the pump was not in use. Approximately \SI{0.2}{s} later, the propellers stopped, and the robot went into free fall.}

In \textbf{Perch State}, \add{assuming the human actively aligns their arm}, the robot performs a free fall and perches on the human arm by \add{rapidly} inflating the joint inflatable actuator to its maximum pressure.
\add{The result, as illustrated in \figref{human_perching}B, indicates that the robot descended \SI{0.11}{m} over \SI{0.3}{s} during arm deformation in midair.
As it fell, the arm deformed due to gravity, caused the joint actuator pressure to momentarily drop to \SI{6}{kPa}, but it recovered to \SI{12}{kPa} at just landing.
In contrast, during the deformation, the bottom actuator pressure increased to \SI{28}{kPa} due to the landing impact force and to \SI{25}{kPa} at landing.}

\add{Following landing, both the pump and bottom actuator supply air to the joint actuators until their pressures equalize, after which only the pump continues inflation.}
As shown in \figref{human_perching}B, both actuators reached \SI{22}{kPa} at \SI{0.4}{s} after landing.
The joint actuator pressure then increased to \SI{30}{kPa} at \SI{0.9}{s}, \SI{40}{kPa} at \SI{1.7}{s}, and \SI{50}{kPa} at \SI{2.6}{s}\extraadd{, corresponding to average rates of \SI{11}{kPa/s} with the pump alone and \SI{25}{kPa/s} when assisted by the bottom actuators.}
When the joint actuator reached \SI{50}{kPa}, the bottom actuator pressure slightly increased to \SI{24}{kPa}, likely due to the increased grasping force from the joint actuator.

\add{The robot successfully perched on a human arm for about \SI{12}{s}, with the joint actuator maintaining an average pressure of \SI{47}{kPa} and the bottom actuator around \SI{22}{kPa}.
During perching, the joint pressure gradually dropped by \SI{6}{kPa}, likely due to minor air leakage from slight damage. 
Although the system is designed to activate the pump if the joint pressure falls below \SI{40}{kPa} or the bottom pressure drops below \SI{20}{kPa}, the pneumatic
system remained off throughout the experiment, indicating that the robot could maintain its grasp without additional energy consumption.}
\add{Moreover, the robot withstood external disturbances including shaking \SI{6.6}{m/s^2} in the z-direction and a \SI{0.45}{rad} roll tilt, while maintaining a stable grasp as shown in \figref{human_perching}A(P-2).}

\add{The pressure changes in the joint and bottom actuators demonstrate that the proposed pneumatic system enables effectively balances rapid arm deformation and optimal pressure for stable grasping while minimizing energy consumption.
The detailed results of the human perching experiment with different bottom inflatable actuator pressures are presented in \appref{different_perching _experiment}.}

\revise{In \textbf{Deperch State}, the robot activates its rotors to take off from the human arm.}
The joint inflatable actuator's air is exhausted by closing the valve connecting it to the bottom actuator and open the exhaust port, while the rotors generate sufficient thrust for instant takeoff.
As illustrated in \figref{human_perching}B, immediately after takeoff, \add{the joint actuator pressure was \SI{13}{kPa}}. 
The residual air prevented the flexible arms from fully straightening, causing an initial descent of about \SI{0.3}{m} in the z-direction and resulting in temporary instability, as illustrated in \figref{human_perching} A(D-2).
However, the robot could quickly recover balance in its posture.
\add{The remaining air was expelled by the thrust as the arms straightened, and the joint actuator pressure eventually reached \SI{0}{kPa}.
The bottom actuator pressure dropped by about \SI{3}{kPa} during take-off due to the loss of grasping force. After returning to normal flight,  the controller resumed maintaining \SI{20}{kPa} as in the approach state.
In the experiment, the human arm was not positioned beneath the robot during takeoff. However, as demonstrated in the object perching experiment, a more stable takeoff can be expected when a supporting surface remains under the robot such as the human arm during takeoff.}

\revise{Despite the close distance between the aerial robot and the human within \SI{0.3}{m}}, neither the robot nor the human was injured, owing to the protective components and bottom inflatable actuator.
Consequently, the effectiveness of the proposed \revise{design, control, and \add{task} planning method} for perching-based human-robot interaction in this study was demonstrated.

\begin{table}[h!]
\centering
\begin{tabular}{c|c|c|c}
\toprule
\textbf{State} &   \textbf{Event/Time(s)}      & \textbf{Joint pressure(kPa)} & \textbf{Bottom pressure(kPa)} \\ \hline\hline
\textbf{Approach} &                            & 0                   & 20               \\\hline
\multirow{2}{*}{\textbf{Reach}}&               & 0                   & 40               \\ \cline{2-4}
 & \textbf{Joint inflated until arm begin to deform}               & 19                  & 27               \\\hline
\multirow{7}{*}{\textbf{Perch}}
                         & \textbf{Free-fall}           & 6                   & 28               \\\cline{2-4}
                         & \textbf{Land}(0)             & 12                  & 25               \\\cline{2-4}
                         & 0.4                 & 22                  & 22               \\\cline{2-4}
                         & 0.9                 & 30                  & 23               \\\cline{2-4}
                         & 1.7                 & 40                  & 23               \\\cline{2-4}
                         & 2.6                 & 50                  & 24               \\\cline{2-4}
                         & 12.6                & 44                  & 22               \\ \hline
\multirow{2}{*}{\textbf{Deperch}} 
                        & \textbf{Takeoff}              & 13                  & 20               \\\cline{2-4}
                         &                     & 0                   & 20 \\
\bottomrule
\end{tabular}
\caption{Pressure Changes in Joint and Bottom Inflatable Actuators in Each State}
    \label{pressure_table}
\end{table}

\section{CONCLUSIONS AND FUTURE WORK} \label{conclusion}
In this paper, we presented the aerial robot can perch on humans to expand the potential applications of aerial robotics in daily living spaces.
By incorporating two key mechanisms, unilateral flexible arms and inflatable actuators, we developed arms that combine rigidity during flight with flexibility for grasping, \extraadd{demonstrating} safe and comfortable interaction with humans.
The pneumatic system enables rapid arm deformation and improves energy efficiency, enhancing grasping performance and perching stability.
\extraadd{Even with the use of standard flight control, robustness against arm deformation during flight is guaranteed, maintaining stable flight.}
\add{The prototype was evaluated for its ability to grasp objects, fit and cling to moving objects and humans.}
Successful in-flight perching on both stationary objects and humans, enabled by task planning for interaction, validates the proposed approach and opens new possibilities for human–aerial-robot interaction.

\add{In future work, object recognition will be integrated to allow the robot to detect outstretched human arms and perch on moving targets, such as walking individuals, enabling more natural and dynamic human interactions. 
Additionally, to enhance grasping stability, the bottom inflatable actuator will be refined to optimize grasping force by automatically adjusting air pressure based on the target.
These advancements will enhance the adaptability for human perching and its practicality in the real-world applications.}

\bibliographystyle{ieeetr}
\bibliography{article.bib}
\section*{APPENDIX}
\renewcommand{\thesubsection}{\Alph{subsection}}

\subsection{Experimental determination of the relationship between air pressure and torque}\label{joint torque experiment}
\add{
Based on the measured data, the coefficients $k_0$, $k_1$, and $k_2$ in \equref{eq:torque} were determined. 
Using these coefficients and the actual design parameters, theoretical values were calculated and validated against experimental results at various pressure levels and joint degrees.}

\begin{figure}[h]
    \centering
    \includegraphics[width=1.0\columnwidth]{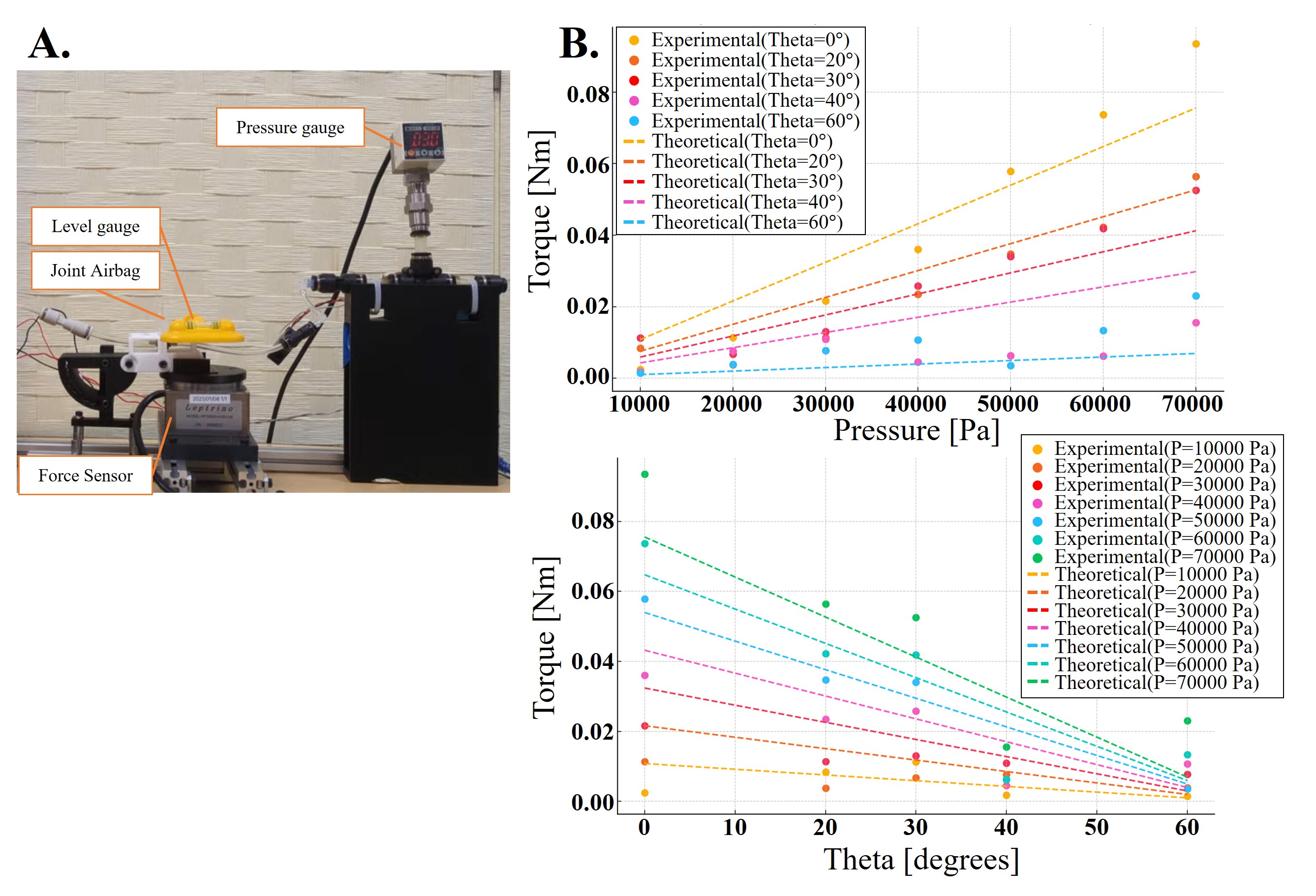}
    \caption{\add{(A) The experimental setup.
    (B) A plot compares the measured and theoretical values. The upper plot represents the relationship between pressure and torque, and the lower plot depicts the relationship between joint angle and torque.}}
    \label{joint_torque}
\end{figure}

\add{In the experiment to obtain the measured data, the air pressure was varied from \SI{0}{} to \SI{70}{kPa} in increments of \SI{10}{kPa}, and the resulting force was measured at joint angles of \( 0 \) \SI{}{rad}, \( \frac{\pi}{9} \) \SI{}{rad}, \( \frac{\pi}{6} \) \SI{}{rad}, \( \frac{2\pi}{9} \) \SI{}{rad}, and \( \frac{\pi}{3} \) \SI{}{rad}. 
The setup for measuring the experimental values are illustrated in \figref{joint_torque}A.
One segment was fixed at a specific angle, while the other was placed horizontally on a force sensor. 
A level gauge was used to ensure the link remains horizontal, and a pressure gauge monitored the air pressure of the joint airbag. 
Using feedback from a pressure sensor, a solenoid valve and pump applied constant pressure to the joint airbag.
The force data from the sensor was averaged over multiple trials and converted to torque by multiplying it by the distance $r =$ \SI{0.0115}{m} from the hinge axis to the center of the segment.}

\add{From the design values of the segment of unilateral flexible arm, the parameters are defined as follows: $y_1 = 0.018 \, \text{m}$, $y_0 = 0.006 \, \text{m}$, $l_{\text{link}} = 0.027 \, \text{m}$
By substituting the values into \equref{eq:torque}, the torque $M$ is expressed as:
\begin{equation}
    M \approx 3.888 \times 10^{-6} (k_0 P_0 - k_1 P_0 \theta) - 5.054 \times 10^{-8} k_2 \theta \label{torque_value}
\end{equation}
To determine $k_0$, $k_1$, and $k_2$, the effects of noise and nonlinearity were considered. The data at the intermediate pressure value of $P =$ \SI{40}{kPa} was selected to minimize these effects. Curve fitting was performed based on the measured data at \SI{40}{kPa}, yielding the coefficients $k_0 \approx 0.2206$, $k_1 \approx 0.1745$, and $k_2 \approx -1.457$. These values were then substituted into \equref{torque_value} and validate it against experimental results.}

\add{A comparison of the measured and theoretical values is shown in \figref{joint_torque}B.
The results indicate that the theoretical predictions follow the trend of the experimental values, suggesting that higher internal pressure in the airbag generates greater torque, whereas at larger hinge angles, its effect on the hinge is reduced. However, deviations between the theoretical and experimental values are observed at higher pressures and larger hinge angles. This deviation is attributed to the influence of measurement noise and the limitations of the current model, suggesting that further refinements or additional correction terms may enhance the model’s accuracy.}

\add{In conclusion, the proposed torque model effectively captures the influence of hinge angle and air pressure, confirming its validity in describing the torque generation mechanism of the airbag to a certain extent.}

\subsection{Experiment on the Influence of the Bottom Inflatable Actuator on Arm Deformation Speed}
\subsubsection{Bottom Inflatable actuator Acts as a Pump for Joint Airbags} \label{bottom faster experiment}
\add{Based on the relationship between the air pressure in the joint actuator and the torque at the arm’s hinge, as presented in \secref{design_req}, we infer that increasing the joint actuator pressure leads to greater arm deformation.  
Therefore, to evaluate the effect of the bottom inflatable actuator on arm deformation speed, we measured the average inflation time over three trials for the joint inflatable actuators in all four arms to inflate from \SI{0}{kPa} to \SI{40}{kPa} under three different conditions:  
\begin{enumerate}
    \item When the pump is directly connected to the joint actuators
    \item When the configuration in (1) incorporates the bottom actuators pre-inflated to \SI{20}{kPa}.
    \item When the configuration in (1) incorporates the bottom actuators pre-inflated to \SI{40}{kPa}.
\end{enumerate}
In the experiments, the four unilateral flexible arms deformed in the direction opposite to gravity, ensuring that bending is driven solely by the joint inflatable actuators for controlled actuation.
The pump output was the same for all scenarios and air from the bottom actuators flowed into the joint actuators as well as air from the pump in the (2) and (3) scenario.}

\add{The results, summarized in \tabref{bending_times}, indicate a reduction in inflation time by \SI{0.37}{s} between conditions (1) and (2), and by \SI{1.03}{s} between conditions (1) and (3). 
The corresponding inflation speeds are calculated as \SI{9.07}{kPa/s} for the pump alone, increasing to \SI{9.90}{kPa/s} in the (2) scenario, and further improving to \SI{11.83}{kPa/s} in the (3) scenario. 
These results reveal that the bottom actuator pre-inflated to \SI{40}{kPa} contributed up to \SI{2.76}{kPa/s} to the overall inflation speed.
}

\add{In conclusion, the experiment demonstrate that the bottom inflatable actuator facilitates faster arm deformation by supplying additional air to the joint actuator. Considering the effects of gravity and external forces acting on the bottom airbags during collisions, the actual arm deformation speed in real-world scenarios is expected to be even faster.}

\begin{table}[h!]
    \centering
    \begin{tabular}{l|c}
        \toprule
        \textbf{Configuration} &  \textbf{time(s)}\\ \hline\hline
        \textbf{Pump Only} & 4.41 \\
        \textbf{Pump + Pre-inflated Bottom Actuator (20kPa)} & 4.04 \\
        \textbf{Pump + Pre-inflated Bottom Actuator (40kPa)} & 3.38 \\ \bottomrule
    \end{tabular}
    \caption{Average Time for the Arm Deforming under Different Configurations}
    \label{bending_times}
\end{table}

\subsubsection{Bottom Inflatable Actuator Adds Additional Torque to the Joint} \label{bottom torque experiment}
\add{To evaluate the function of the torque generated by the bottom actuator in enhancing the arm deformation speed, we compare the average time over three trials from the moment the joint inflatable actuator begins pressurization until the single unilateral flexible arm bends to \( \frac{\pi}{4} \), \( \frac{\pi}{2} \), or \( \frac{3\pi}{4} \) \SI{}{rad}.  
The comparison was made between two conditions: where a pressurized bottom airbag at \SI{0}{kPa}, \SI{20}{kPa}, or \SI{40}{kPa} was fixed to the two links, and one where it was not fixed. } 

\add{In this experiment, the unilateral flexible arm was equipped with a motion capture system and deformed in the direction opposite to gravity, ensuring that the deformation was driven solely by the joint inflatable actuator rather than by gravitational force. 
The pump output remained constant across all conditions. When the bottom airbag was pre-inflated to \SI{20}{kPa} or \SI{40}{kPa}, air from both the pump and the bottom airbag flowed into the joint actuator.}

\add{The results are summarized in \tabref{arm_bending}.
First, in the absence of torque, increasing the bottom actuator pressure from \SI{0}{kPa} to \SI{40}{kPa} led to longer deformation times across all angles. This suggests that the expansion of the bottom actuator hinders arm deformation, and at higher pressures, interference between inflated sections further contributes to this effect.}

\add{However, when comparing conditions with and without torque, deformation time was reduced at \SI{20}{kPa} and \SI{40}{kPa}, whereas at \SI{0}{kPa}, no significant difference was observed. Additionally, higher pressurization resulted in a greater reduction in deformation time. This effect was particularly pronounced when the bottom actuator was pressurized to \SI{40}{kPa}, where at \( \frac{3\pi}{4} \), the deformation time was reduced by \SI{0.44}{s} compared to the condition without torque.  }

\add{The shortest average arm deformation time was observed at \( \frac{\pi}{4} \) and \( \frac{\pi}{2} \) when torque was applied and the bottom actuator was pressurized to \SI{40}{kPa}. In contrast, at \( \frac{3\pi}{4} \), the shortest deformation time occurred both when torque was applied with \SI{40}{kPa} pressurization and when the bottom actuator was unpressurized at \SI{0}{kPa}.  }

\add{In conclusion, these results indicate that the inhibitory effect of bottom actuator pressurization on arm deformation can be counteracted by applying torque. Furthermore, since this study focuses on using torque to accelerate arm deformation at its initial stage, the experiment successfully demonstrated the effectiveness of this approach in facilitating faster deformation.}

\begin{table}[h!]
    \centering 
    \begin{tabular}{c|@{}c|cc@{}}
    \toprule
    \textbf{Arm Deformation Degree} &\textbf{Bottom Pressure (kPa)} & \textbf{Without Torque (s)} & \textbf{With Torque (s)} \\ \hline\hline
    \multirow{3}{*}{\textbf{\(\frac{\pi}{4}\)}} 
    &0 & 1.13    & 1.14     \\
    &20 & 1.21   & 1.19    \\
    &40 & 1.23   & 1.07    \\ \hline
    \multirow{3}{*}{\textbf{\(\frac{\pi}{2}\)}} 
    &0 & 1.24    & 1.25     \\
    &20 & 1.31   & 1.30    \\
    &40 & 1.43   & 1.20    \\ \hline
    \multirow{3}{*}{\textbf{\(\frac{3\pi}{4}\)}} 
    &0 & 1.34    & 1.36     \\
    &20 & 1.45   & 1.43    \\
    &40 & 1.78   & 1.34    \\
    \bottomrule
    \end{tabular}
    \caption{Comparison of Arm Deforming Time with and without Torque}
    \label{arm_bending}
\end{table}

\subsection{Theoretical and Experimental Analysis of Inflatable Actuator Pressure and Volume}\label{pneumatic equation}
\begin{figure}[h!]
    \centering
    \includegraphics[width=0.9\columnwidth]{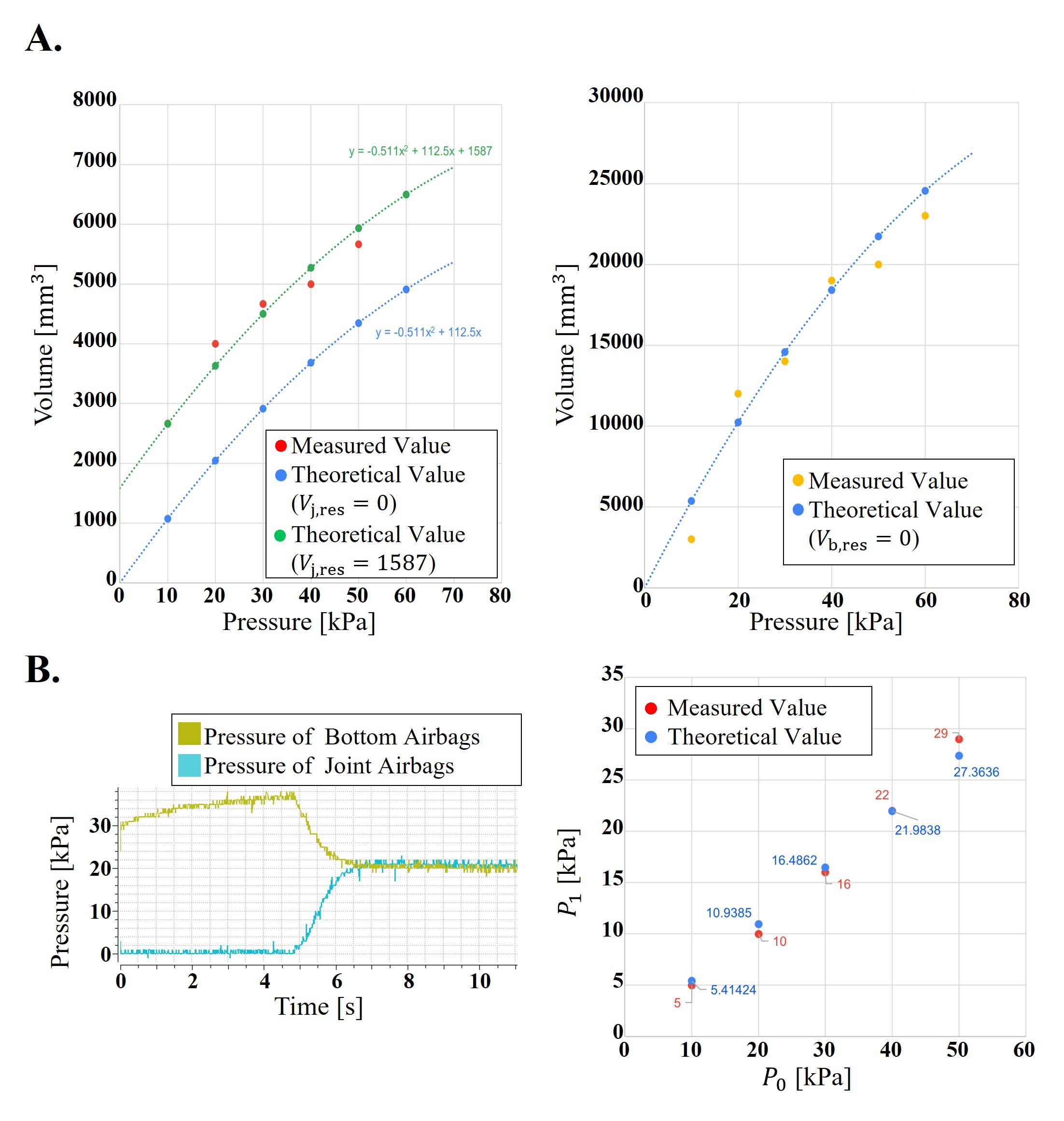}
    \caption{
      (A) Theoretical and measured volumes of the joint and bottom inflatable actuators
      (B) Theoretical and measured values of the initial bottom pressure \( P_0 \) and the stabilized pressure \( P_1 \). The plot on the left shows the pressures of the joint and bottom inflatable actuators during an experiment where the initial bottom pressure \( P_0 =\) \SI{38}{kPa}.
      }
    \label{P_0_and_P_1}
\end{figure}

\add{Based on the design of each inflatable actuators, the parameters are set as follows:  
\( x_j= 30\) \SI{}{mm}, \( y_j = 20\)\SI{}{mm}, \( x_{b,1} = 20\)\SI{}{mm}, \( x_{b,2} = 15\)\SI{}{mm}, \( x_{b,3} = 10\)\SI{}{mm}, \( x_{b,4} = 30\)\SI{}{mm}, \( x_{b,5} = 25\)\SI{}{mm}, \( y_b = 40\)\SI{}{mm}, \( n_b = 4 \), and \( n_j = 20 \).  
For the joint inflatable actuators, the long side deforms more than the short side due to the folds when pressurized. Therefore, we set \( x_j = 30\)\SI{}{mm} and \( y_j = 20\)\SI{}{mm}.  
Additionally, since empirical observations show that the inflatable actuators collapses at \SI{80}{kPa}, we set \( P_{\max} = 80 \).  
By substituting these values, \equref{final equ P1} is expressed as follows:
\begin{align}
    V_{b0} &\approx \sum_{i=1}^{4}\left(-2.555 P_0^2+562.5 P_0 + V_{\text{b,res}}\right) \label{V_b0_app}\\ 
    V_{b1} &\approx \sum_{i=1}^{4}\left(-2.555 P_1^2+562.5 P_1 +V_{\text{b,res}}\right) \label{V_b1_app}\\
    V_{j0} &= \sum_{i=1}^{20}{V_{\text{j,res}}} \\
    V_{j1} &\approx \sum_{i=1}^{20}\left(-0.511 P_1^2 +112.5 P_1+ V_{\text{j,res}}\right) \label{V_j1_app}
\end{align}
The values of \( V_{\text{b,res}} \) and \( V_{\text{j,res}} \) are determined empirically.
We use Archimedes' principle by submerging the airbag in water, pressurizing it, and measuring the weight of the displaced water to estimate its volume.
For the joint airbags, we measure the volume of three samples and use their average value.  
We apply function fitting to the measured values and obtain \( V_{\text{b,res}} =\) \SI{0}{mm^3},\( V_{\text{j,res}} =\) \SI{1587}{mm^3}.   
\figref{P_0_and_P_1}A shows the comparison between the measured and theoretical volumes.  
For the joint inflatable actuator, the approximation error ranges from \SI{-367}{mm^3}(\SI{30}{kPa}) to \SI{269}{mm^3}(\SI{50}{kPa}), while for the bottom inflatable actuator, the error ranges from \SI{-588}{mm^3}(\SI{40}{kPa}) to \SI{2369}{mm^3}(\SI{10}{kPa}).}
\add{Substituting \( P_{\text{atm}} \approx\) \SI{101.3}{kPa}, \( V_{\text{b,res}} \) and \( V_{\text{j,res}} \), \equref{P_1} can be written as followed:
\begin{align}
    &(P_1 + 101.3) \left( -20.44 P_1^2 + 4500 P_1 \right) \notag \\
    &\approx(P_0 + 101.3) \left( -10.22 P_0^2 + 2250 P_0 \right) \notag \\
    &\quad+ 31740(P_0 - P_1)
\end{align}}
\add{The comparison between theoretical and actual values for the relationship between \( P_0 \) and \( P_1 \), as shown in \figref{P_0_and_P_1}B.}
\add{In the experiment, when the initial bottom pressure \( P_0 \) was set to \SI{38}{kPa}, the stabilized pressure \( P_1 \) reached \SI{21}{kPa}, which is close to the theoretical value. 
Additionally, even when \( P_0 \) was varied, the theoretical and measured values appear to show minimal differences.}

\subsection{Fatigue Test of Airbag} \label{fatigue_test}
\add{To evaluate the durability of the joint inflatable actuator, a fatigue test was conducted by repeatedly inflating and deflating the actuator.
In this test, air was supplied until the internal pressure reached a predetermined threshold, at which point the pressure was released to \SI{0}{kPa}. This cycle was repeated until the actuator could no longer reach the threshold pressure, and the number of cycles until failure was recorded.
However, since the experiment was conducted using a single airbag, the effect of the time lag in sensor feedback to the pump became significant. As a result, when the threshold pressure was set to \SI{50}{kPa}, the actual pressure fluctuated between \SI{50}{kPa} and \SI{57}{kPa}, whereas at a \SI{40}{kPa} threshold, the pressure varied between \SI{40}{kPa} and \SI{50}{kPa}.}

\add{The test results showed that when the threshold was set to \SI{50}{kPa}, the actuator failed after an average of 289 cycles in three trials, with a maximum of 428 cycles. In contrast, when the threshold was set to \SI{40}{kPa}, the actuator withstood an average of 2144 cycles in three trials, with a maximum of 3275 cycles.
These results indicate that lowering the inflation pressure significantly improves the durability of the joint inflatable actuator.}

\subsection{\revise{Autonomous Human Perching Behavior}} \label{planning}
\begin{figure}[h!]
    \centering
    \includegraphics[width=1.0\columnwidth]{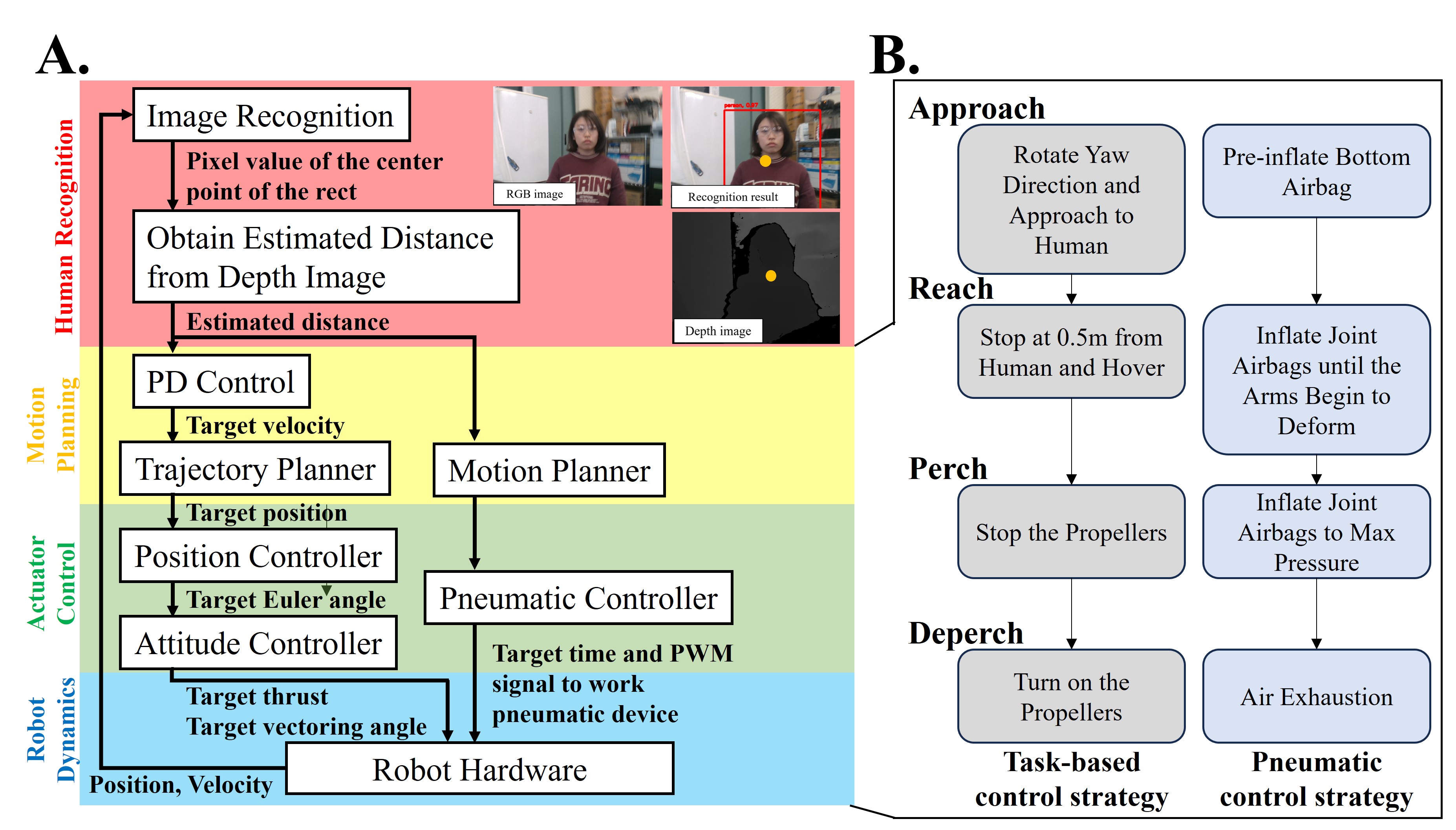}
    \caption{\add{Task-based control strategy} of the aerial robot that can perch on humans.
      (A) System architecture
      (B) Finite state machine of this perching system}
    \label{recog_system}
\end{figure}

\revise{
By getting close to the person who is the target to perch, we believe that instead of the robot perching on its own, the robot can induce the person to attract interest and actively extend his/her arm like a falconer so that the robot can hold onto it. 
Therefore, we develop recognition and \add{task} planning methods according to such behavior strategy as shown in \figref{recog_system}.
}

\subsubsection{\revise{Motion Sequence} of Perching on Human's Arm}
As illustrated in \figref{recog_system}B, the robot \revise{follows} a sequence of actions toward humans \revise{with} five states: Search, Approach, Reach, Perch, and Deperch. 

\begin{description}
    \item[\textbf{Approach}]~\\
    The approach motion to the vicinity of the target person involves both rotational and translational movements based on human recognition. For rotational movement, the robot is required to modify the yaw angle to face the person.
    Human recognition is performed based on onboard sensor that provides both RGB image and the depth image for detecting the human location in a three-dimensional manner as illustrated in \figref{recog_system}A.
    A deep learning-based detection method \add{\cite{howard2017mobilenets,tan2020efficientdet} by edge AI accelerator} is applied to estimate the bounding box and the center point of the target person, which is subsequently used to obtain the distance from the corresponding depth image. 
    In cases where multiple individuals are detected, the system prioritizes the closest person with the largest bounding box area.

    For translational movement, to ensure the safety of human-aerial robot interaction, we propose a \revise{velocity-based control} to approach the target \revise{person} in which the robot moves slower as it gets closer to the person.
    The distance between the robot and the target, which is estimated from the depth image, is used as input to obtain the desired velocity to approach the person by \revise{following} PD control:
    \begin{align}
      v^{des} &=  K_ {p} d^{\text{err}} + K_{d} \dot{d}^{\text{err}}, \label{eq:chap5:PID}\\
      d^{\text{err}} &= d^{\text{des}} - d.
    \end{align}
    \add{The depth threshold is continuously applied to exclude outliers. When the depth values is unavailable, the robot temporarily stop, if the exceeded the threshold, the previous value is used instead.}
    The robot approach's goal is \SI{0.5}{m} (that is $d^{\text{de}}$) away from the human to elicit the human's instinctive movement to extend his/her arm and touch each other.
    \add{Additionally, the bottom inflatable actuator is inflated to some extend to absorb impact in case of an unexpected fall.}
    
    \item[\textbf{Reach}]~\\
    The robot hovers in place and prepares for perching while the human actively extends their arms under the robot.
    This preparation time is designed to be approximately the same time it takes the falconer to perch the falcon on his arm.
    \add{During hovering, the bottom inflatable actuator expand to \(P_0\) as stated in \secref{determine_p} to prepare for rapidly expanding joint inflatable actuator during arm deformation.
    When the bottom actuator reaches the preset pressure, the valve opens, allowing air to flow into the joint actuators.
    Once the joint inflatable mechanism inflates to the maximum pressure that maintains arm rigidity, the propellers stop, and the robot enters free fall.}
    
    \item[\textbf{Perch}]~\\
    The propellers are shut off, and the robot free-falls towards the human's arm below. 
    At the same time, by \add{flowing air from the maximum output Pump and bottom inflatable actuators into the joint inflatable actuators}, the arms deform \add{rapidly} to grasp the human's arm, enabling perching.
    \add{For successful perching, the human user must proactively align their arm with the robot, as the robot does not actively recognize the arm. This design is intended to encourage deliberate user engagement in the interaction process.}
    
    \item[\textbf{Deperch}]~\\
    It takes off from the human arm, called "Deperch."
    The propellers generate thrust for flight while simultaneously opening the exhaust ports to reduce the internal pressure of the \add{joint} inflatable actuators. This allows the arms to straighten instantly, and the robot lifts off from the arm linearly. 
\end{description}

\subsection{Evaluation of the Grasping Force Affected by Joint and Bottom Inflatable Actuator}\label{grasping}
\add{To evaluate the influence of joint and bottom inflatable actuators on grasping force, we conduct experiments measuring the grasping force when grasping objects of multiple shapes and sizes.
As shown in \figref{evaluation_grasping_force}B-1, we use an arm attached to a force sensor to grasp an object in the experiments. A weight (\SI{0}{}-\SI{2.5}{kg}) and a plastic bottle were placed in the center of the object, and the weight gradually increases by increasing weight or adjusting the amount of water. 
The maximum grasping force is calculated by subtracting the baseline force in the unloaded state from the force at the moment the object begins to slip. The tested objects are cylindrical objects with diameters of \SI{55}{mm} and \SI{100}{mm}, an arm model based on the average arm dimensions data of the adult male\cite{AISTdatabase}, and rectangular objects with base dimensions of \SI{110}{mm} × \SI{110}{mm} and heights of \SI{50}{mm} and \SI{100}{mm}.
The plot diagrams for all experiments are shown in \figref{evaluation_grasping_force}A.
}

\subsubsection{Grasping Posture Configuration}
\add{We compare the grasping force of cylindrical objects of \SI{55}{mm} and \SI{100}{mm} diameter and the arm model using 4 and 2 arms. The results consistently show that the grasping force is higher with the four-arm configuration for all objects. 
Thus, in this study, we employed a four-arm grasp configuration.}

\subsection{Comparison with Perching and Grasping Mechanisms} \label{comparison}
\add{We present \tabref{grasping_mechanisms} summarizing the key features of two categories of perching mechanisms, including the arm material, weight, driving method, energy consumption to transition to each state, and maximum grasping force.}

\add{We compare these mechanism based on categories.
Firstly, we compare the independent grasping mechanisms and the robot in this study in terms of perching on a human.
Robots such as Ultra Hand\cite{mclaren2019passive}, HFB Gripper\cite{nguyen2023soft}, and SNAG Leg\cite{roderick2021bird} utilize passive grasping mechanisms to activate their arms during perching by releasing energy stored in the springs. This approach minimizes energy consumption and allows for rapid perching. In particularly, HFB Gripper\cite{mclaren2019passive} and SNAG Leg\cite{roderick2021bird} are triggered by the impact force during perching. 
However, they cannot control the grasping force during activation. Therefore, grippers that can generate large grasping force, such as Ultra Hand\cite{mclaren2019passive} and HFB Gripper\cite{nguyen2023soft} are not suitable for human arms because humans may feel pain due to the large uncontrolled grasping force and impact force on the human arm, although specific impact force values are not stated.
}
\add{Furthermore, after perching, which we consider the most significant issue of the independent grasping mechanism, the flight mechanisms above the grasping mechanism such as SoBAR(size: \SI{319}{}×\SI{319}{mm})\cite{nguyen2023soft}, SNAG(overall size: \SI{500}{mm})\cite{roderick2021bird} and the aerial robot mounted on Ultra Hand(size: \SI{319}{}×\SI{319}{mm})\cite{mclaren2019passive} are too large and could restrict the movement of the human arm during grasping. 
For example, since the average forearm length for a Japanese adult is about \SI{240}{mm}\cite{sizeJPN2004_2006} and the average elbow-wrist length for a American adult male is about \SI{290}{mm}\cite{sizeUSA2016}, it would be difficult for them to bend their elbow while these robots are perched on their forearms.
In contrast, the aerial robot in this study has the great advantage of not interfering with the natural movement of the human arm because it can be reduced to a compact size of\SI{110}{} × {110}{mm} above and {290}{} × {290}{mm} below the human arm when grasped.
Additionally, to accommodate such movements, a minimum grasping force of \SI{20}{kPa} is considered necessary. This requirement makes it difficult for designs such as the Compliant Bistable Gripper\cite{8793936} or the Soft Gripper\cite{ubellacker2024high} to respond effectively to such demands.}

\add{Regarding the weight, compared to existing studies on deformable robots that perform grasping, the weight of the robot developed in this study is the second lightest, following Metamorphic Aerial Robot\cite{zheng2023metamorphic}. Among comparable metrics, it demonstrates the best performance in both grasping force and energy efficiency.
}
\add{Focusing on energy consumption during arm holding, our proposed robot achieves the highest energy efficiency in grasping among the comparable designs. If no additional air is supplied during perching, it requires minimal power to maintain its grasping force. 
As a result, among the evaluated designs, the ratio of the grasping force to power consumption is significantly higher than that of a Soft Gripper\cite{ubellacker2024high}  or SNAG\cite{roderick2021bird}, making it the most efficient system in terms of energy usage.}

\add{Furthermore, compared to pneumatically driven soft grippers\cite{9828332,10521918}, which enable grasping force control, the proposed robot’s arm deformation time, pressure for grasping, weight, and max grasping force are summarized in \tabref{soft_gripper}. The proposed robot operates at a lower pressure while achieving a higher grasping force than existing pneumatically actuated soft grippers, enhancing energy efficiency. Additionally, the bottom inflatable actuator functions as an air reservoir, allowing for a shorter actuation time required for arm deformation.
}

\begin{table}[h!]
    \centering
    \renewcommand{\arraystretch}{0.95} 
    \begin{tabular}{p{2.0cm}|p{2.0cm}|p{2.2cm}|p{1.2cm}|p{1.9cm}|p{1.9cm}|p{1.9cm}|p{1cm}}
    \toprule
    \textbf{Category} & \textbf{Name (Robot Name)} & \textbf{Material} & \textbf{Weight\newline(kg)}& \multicolumn{3}{c|}{\textbf{Drive Method [Energy Consumption]}} & \textbf{Max Grasp Force (N)} \\ \cline{5-7}
    &&&&\textbf{Activate} & \textbf{Hold} & \textbf{Retract}&\\ \hline\hline
    \multirow{5}{3cm}{Independent \\ Grasping \\ Modules}
        & Ultra Hand\cite{mclaren2019passive} & Polyurethane elastomer
            & 0.51 
            & Passive by spring energy \newline [-]& Passive \newline [-]& Tendon \newline [-]
            & 56 \\ \cline{2-8}
        & Soft Gripper \cite{ubellacker2024high} & Expanding polyurethane foam  
            & 0.54 
            & Passive by form elastance \newline [0.29W]& Passive\newline [0.01W] & Tendon\newline [-] 
            & 19.6 \newline(2kg) \\ \cline{2-8}
        & SNAG Leg \newline (SNAG) \cite{roderick2021bird} & Hard plastic 
            & 0.25 
            & Passive by spring energy \newline [1.8W\textsuperscript{[1]}]& Passive \newline [0.10W\textsuperscript{[2]}] & Tendon \newline [-] 
            & 4.9 \newline(0.5kg) \\ \cline{2-8}
        & HFB Gripper \newline (SoBAR)\cite{nguyen2023soft} & Hybrid fabric with spring steel 
            & 0.30\textsuperscript{[3]} 
            & Passive by tape spring energy\newline [-] & Passive \newline [-] & Pneumatic \newline (83kPa\textasciitilde) \newline [-]
            & 200 \\ \cline{2-8}
        & Compliant Bistable Gripper\cite{8793936} & Polyurethane elastomer 
            & 0.036 
            & Passive by contact force\newline [0W\textsuperscript{[4]}]& Passive \newline [0W\textsuperscript{[4]}]&Heat \newline [-]& 0.6 \\ \midrule
    \multirow{4}{3cm}{Deformable \\ Bodies}
        & SOPHIE\cite{9851515} & Flexible TPU 
            & 1.8 
            & Tendon \newline [-]& Tendon \newline [30.27W\textsuperscript{[5]}] & Tendon \newline [-]
            & - \\ \cline{2-8}
        & Metamor-\newline phic Aerial Robot\cite{zheng2023metamorphic} 
        & Sandwich of CFRP, polyimide sheet, \newline elastic latex membrane  
        & 0.65\textsuperscript{[6]}
        & Tendon \newline [-] & Tendon \newline [0.48W\textsuperscript{[7]}\textasciitilde] & Tendon \newline [-] 
        & - \\ \cline{2-8}
        & WHOPPEr\cite{tao2023design} & 3D-printed links with 3-layer bistable tape spring 
        & 1.6 
        & Tendon \newline [-] &Passive \newline [-] &Thrust \newline(+Tendon) \newline [-]
        & 40 \\ \cline{2-8}
        & This study & 3D-printed links with airbags 
        & 1.444 
        & Pneumatic (\textasciitilde50kPa) \newline [0.17\textasciitilde \newline4.03W\textsuperscript{[8]}] & Pneumatic \newline [0.17W\textasciitilde] & Pneumatic \newline [0.50W]\newline +Thrust 
        & 45.2 \\ \bottomrule
\end{tabular}
\raggedright
\scriptsize{
\textsuperscript{[1]} This estimated power consumption is only for the foot motors from operating at 6V and drawing 0.3A of current. 
\textsuperscript{[2]} Estimated based on the grasping mechanism run on 5V and the resting current for the grasping mechanism is 0.019 A. 
\textsuperscript{[3]} Calculated from the weight of the gripping mechanism and pump (NMP830 HP-KPDC-B). 
\textsuperscript{[4]} Estimated to have no standby power because the mechanism is driven by the contact force with the object.
\textsuperscript{[5]} Estimated based on the battery capacity (2S 1200mAh) and the maximum operating time of 17.6 minutes during perching. 
\textsuperscript{[6]} The weight without battery. 
\textsuperscript{[7]} Estimated from the state that the servomotor(Dynamxiel XH430-W350) consumes 40mA standby current during perching on the smaller branches.
\textsuperscript{[8]}The maximum power consumption is when the pump is operating at maximum capacity.\par
}
\caption{Grasping Mechanisms Overview. The weight refers to the gripper's weight for independent grasping graspers and the total weight of the robot for deformable body}
\label{grasping_mechanisms}
\end{table}

\begin{table}[h!]
    \centering
    \begin{tabular}{c|c|c|c|c|c|c}
        \toprule
        \textbf{Category} & \textbf{Name} 
        & \shortstack{\textbf{Weight} \\ \textbf{(kg)}} &
        \shortstack{\\\textbf{Arm(Finger)} \\\textbf{length (mm)}} &
        \shortstack{\\\textbf{Max Grasping} \\\textbf{Force (N)}} & 
        \shortstack{\textbf{Pressure} \\ \textbf{(kPa)}}
        &\shortstack{\textbf{Deformation} \\ \textbf{Time (s)}} \\ 
        \hline\hline
        \multirow{2}{*}{\shortstack{Independent \\ Grasping \\Modules}}
        & \shortstack{\\Pneumatic \\Soft Gripper 1\cite{9828332}} 
        & 0.55 &110& \shortstack{7.99\\(0.814 kg)} & 110 & - \\ \cline{2-7}
        & \shortstack{\\Pneumatic\\ Soft Gripper 2\cite{10521918}} & 0.808 
        &100& \shortstack{2.13\\(0.217 kg)} & 58 & 5 \\ 
        \midrule
        \shortstack{\\Deformable \\Body} & This study & 1.444 
        & 150 &45.2 & 50 & 3.38\textsuperscript{[1]}(40kPa) \\ 
        \bottomrule
    \end{tabular}

\raggedright
\scriptsize{
\textsuperscript{[1]} The arm morphs in the direction opposite to gravity.
}
    \caption{Comparison of Pneumatic Soft Grasping Mechanisms}
    \label{soft_gripper}
\end{table}

\subsection{\revise{Dynamics Model around the Near-hovering State}} \label{flight_control}
\revise{Give that the flexible arms are rigid in near-hovering state, the general translational and rotational dynamics of the quadrotor can be given as follows:}
\begin{align}
  \revise{m} \ddot{\bm{r}} &= \bm{R}\force - \revise{M} \bm{g}, \label{eq:newton_linear_motion} \\
  \bm{I} \dot{\bm{\omega}} &= \revise{\torque - \bm{\omega} \times \bm{I} \bm{\omega}}, \label{eq:euler_angular_motion}
\end{align}
\revise{
where \revise{$m$}, $\bm{I}$, $\bm{r}$, $\bm{R}$, and $\bm{\omega}$ represent the mass, the inertia tensor, the position vector of the center of mass, the attitude matrix, the angular velocity vector of the robot, respectively, and $\bm{g}$  denotes the gravitational acceleration vector.
}
\revise{Under the assumption that the flexible arm is rigid during hovering flight, the force $\force$ and torque $\torque$ can be further given as follows:}
\begin{align}
  \force &= \sum_{i=1}^4 \ui \lambda_i,
  \label{eq:thrust_to_force} \\
  \torque &= \sum_{i=1}^4 (\p \times \ui + \sigma_i \ui)\lambda_i,
  \label{eq:thrust_to_moment}
\end{align}
\revise{
where $\lambda_i$, $\sigma_i$ denote the thrust force and the drag rate of $i$-th rotor thrust respectively.
$\bm{p}_{i} =[p_{i_x} \: p_{i_y} \: p_{i_z}]^{\top}$ is the the position of the $i$-th rotor as seen from the body coordinate system, while the thrust direction vector $\ui =[{u}_{i_x} \: {u}{i_y} \: {u}_{i_z}]^{\top}$ of the $i$-th rotor as seen from the body coordinate system.
}
\revise{Then}, the wrench $\bm{w}=[f_x \: f_y \: f_z \: \tau_x \: \tau_y \: \tau_z]^{\top}$ is given by:
\begin{align}
 \label{eq:wrench_quad}
  \bm{w}
  &=
  \begin{bmatrix}
     \force \\
     \torque
  \end{bmatrix} 
  =
  \begin{bmatrix}
     \bm{Q}_{trans} \\
     \bm{Q}_{rot}
  \end{bmatrix}  \thrust
  \revise{=}
 \bm{Q} \thrust,
\end{align}
where
\begin{align}
  \bm{Q}_{trans} &=[\bm{u}_1  \cdots \bm{u}_4] ,\\
  \bm{Q}_{rot} &= [\bm{p}_1 \times \bm{u}_1 + \sigma_1  \bm{u}_1  \cdots \bm{p}_4 \times \bm{u}_4 + \sigma_4 \bm{u}_4].
\end{align}

\subsection{\revise{Flight Control}}

\subsubsection{Attitude Control}\label{attitude_control}

Attitude control in this study uses LQI control\cite{young1972approach}, which adds an integral term to \revise{compensate steady error}. Integrating \eqref{eq:euler_angular_motion} and \eqref{eq:wrench_quad},
the state equations for the quadrotor attitude can be represented as follows:
\begin{align}
  \dot{\bm{x}} &= \bm{A}\bm{x} + \bm{B}\bm{\lambda} - \bm{D}(\bm{I}^{-1} \bm{\omega} \times \bm{I} \bm{\omega}), \\
  \bm{y} &= \bm{C}\bm{x},
\end{align}
\revise{
where we use XYZ Euler angles ($\begin{bmatrix} \wphi & \wtheta & \wpsi \end{bmatrix}$) to express the rotation: 
\begin{align}
  \bm{x} =
  \begin{bmatrix}
    \wphi &  \omega_x & \wtheta & \omega_y & \wpsi &  \omega_z
  \end{bmatrix}.
\end{align}
Note that we assume the derivative of Euler angles are identical to $\bm{\omega}$ in near hovering state.
}

The integral term $\bm{v}$ can be expressed as:
\begin{equation}
  \dot{\bm{v}}= \bm{y}- \bm{y^{des}} = \bm{C}(\bm{x} - \bm{x^{des}}) = \bm{Ce}.
  \label{eq:dot_v}
\end{equation}

Thus, \equref{eq:dot_v} can be rewritten as:
\begin{equation}
  \dot{\bar{\bm{x}}} = \bar{\bm{A}}\bar{\bm{x}} + \bar{\bm{B}}\bm{\lambda} - \bar{\bm{D}}(\bm{I}^{-1} \bm{\omega} \times \bm{I} \bm{\omega}),
\end{equation}
where
\begin{align}
  \bar{\bm{x}} &=
  \begin{bmatrix}
    \bm{e} & \bm{v}
  \end{bmatrix}^{\top}, \\
  \bar{\bm{B}} &=
  \begin{bmatrix}
    \bm{0}_{4 \times 1} & \bm{B}_1 &
    \bm{0}_{4 \times 1} & \bm{B}_2 &
    \bm{0}_{4 \times 1} & \bm{B}_3 &
    \bm{0}_{4 \times 3}
  \end{bmatrix}^{\top},
\end{align}
\begin{equation}
  \begin{bmatrix}
    \bm{B}_1 & \bm{B}_2 & \bm{B}_3
  \end{bmatrix}^{\top} = \bm{I}^{-1} \bm{Q}_{rot}.
\end{equation}

The cost function is represented using a diagonal weight matrices $\bm{M}$,$\bm{N}$ as follows:
\begin{equation}
  \bm{J} = \int_0^\infty (\bar{\bm{x}}^{\top} \bm{M}\bar{\bm{x}} + \bm{\lambda}^{\top} \bm{N} \bm{\lambda} ) dt
  \label{eq:cost_function}
\end{equation}

Finally, the feedback gain matrix $\bm{K}$ is derived by solving the Riccati equation, and the input for attitude control is expressed as:
\begin{equation}
\bm{\lambda}^{des}_{rot} = \bm{K} \bm{x} + \bm{Q_{rot}}^{\#} \bm{I}^{-1}  \bm{\omega} \times \bm{I} \bm{\omega}
\end{equation}
\revise{where $\bm{Q_{rot}}^{\#}$ is the Moore–Penrose inverse of matrix $\bm{Q_{rot}}$.}

\subsubsection{Position Control} 
Position control in this study uses general PID control.
Given the desired position as input, the target force $\force^{des}$ is expressed as follows using PID control.
\begin{align}
  \force^{des} &= \revise{m} \bm{R}^{-1}(\bm{g} + \bm{K}_{r,p}  \pos^{err}  \nonumber \\
  & + \bm{K}_{r,i} \int \pos^{err} dt + \bm{K}_{r,d} \dot{\pos}^{err}),
  \label{eq:chap4:desire_force}\\
  \pos^{err} &= \pos^{des} - \pos,
\end{align}
where $\bm{K}_{r,p}$, $\bm{K}_{r,i}$, and $\bm{K}_{r,d}$ are diagonal matrices representing the proportional gain, integral gain, and derivative gain, respectively.
\revise{
For the conversion from the desired force (that is linear acceleration) to the desired attitude, we follow the following rule:}
\begin{align}
  \label{eq:desired_roll}
  \wphi^{des} &= tan^{-1}(-\bar{f}_y, \sqrt{\bar{f}^2_x + \bar{f}^2_z}) \\
  \label{eq:desired_pitch}
  \wtheta^{des} &= tan^{-1}(\bar{f}_x,\bar{f}_z^{\mathrm{des}}) \\
   \left[\bar{f}_x \ \bar{f}_y \ \bar{f}_z \right]^{\top}  &= R_z^{-1}(\wpsi^{des}) {\bm f}^{des} \nonumber
\end{align}
where $R_{Z}(\cdot)$ is a special rotation matrix which rotates only along the $z$ axis.

\subsection{Perching Experiments with Different Bottom Air Pressures} \label{different_perching _experiment}
\begin{figure}[h!]
    \centering
    \includegraphics[width=1.0\columnwidth]{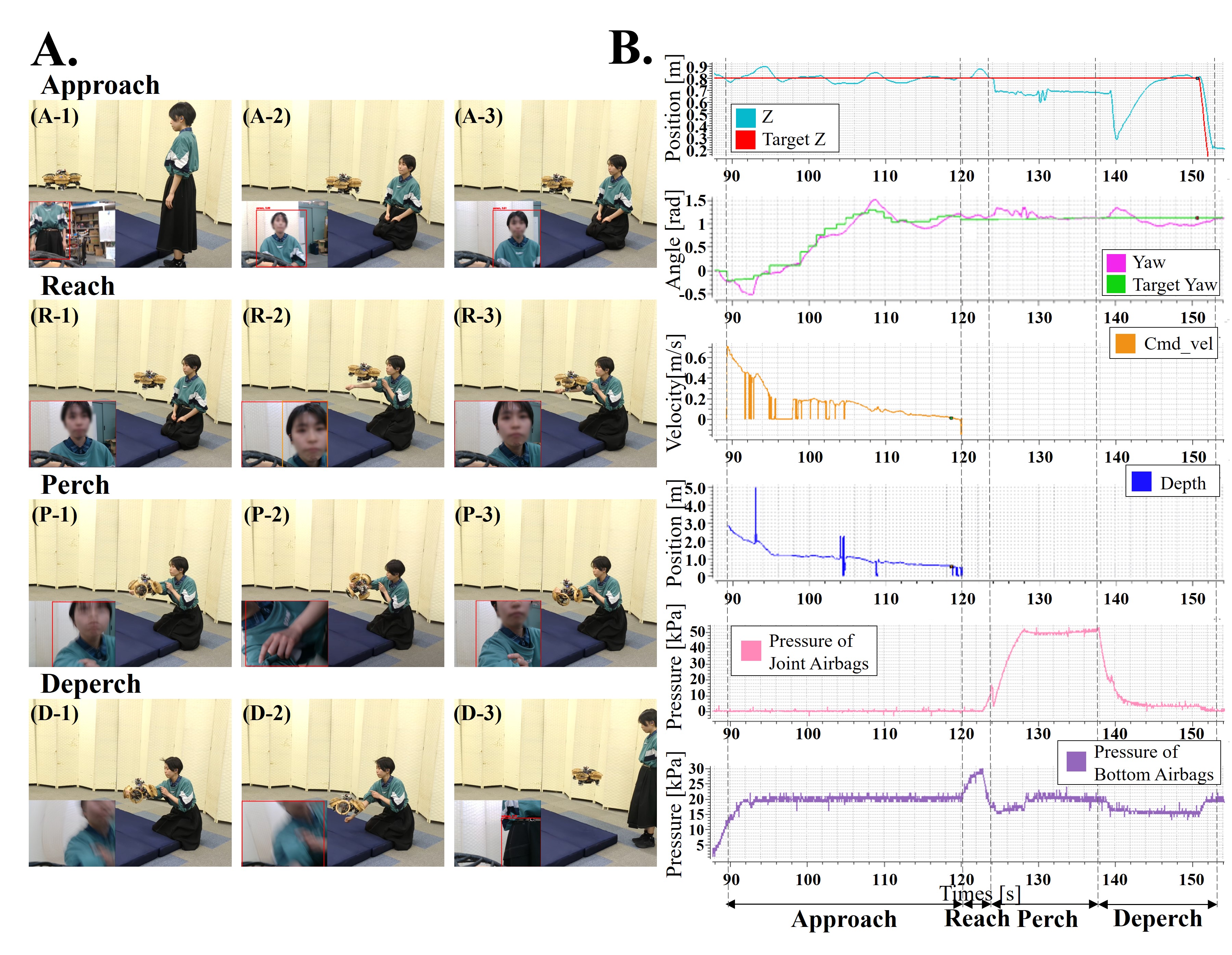}
    \caption{Experiment of perching on a human arm from in-flight situation when the pre-applied pressure of bottom inflatable actuator \SI{30}{kPa}.
    (A) Snapshots of the experiment of perching on a human arm. The small sub-image below the left of each image is the human recognition result using an onboard RGB camera image.
    (B) plots of the target and measured \add{z position and yaw angle, along with the desired velocity and estimated distance towards the target person and the pressure of joint and bottom inflatable actuators}}
    \label{human_perching_30}
\end{figure}

\add{The experiment of bottom actuator's pre-applied pressure \SI{30}{kPa}} is shown in \figref{human_perching_30}.
In the experiment, the graph as shown in \figref{human_perching_30}B, indicates that that \SI{1.6}{s} after the bottom actuator reached \SI{30}{kPa}, both actuators reached \SI{17}{kPa}. Approximately \SI{0.1}{s} later, the propellers stopped, and the robot went into free fall.
\add{Considering that a typical pump has a startup time, the expansion of the joint actuator before free fall was likely primarily driven by the bottom actuator. 
However, since the stabilized pressure in this case exceeded the theoretical value and the experimental value by \SI{1}{kPa}, as detailed in \appref{pneumatic equation}, it is possible that the pump had a slight influence as well.}

\add{In Perch state, the robot descended \SI{0.12}{m} over \SI{0.3}{s} while deforming its arm in midair.
As it fell, the arm deforms due to gravity, caused the joint actuator pressure to momentarily drop to \SI{3}{kPa}, but it recovered to \SI{12}{kPa} at landing. The landing impact force increased the bottom actuator pressure to \SI{25}{kPa}.}

\add{Beside, \SI{0.5}{s} after landing, both actuators reached \SI{}{kPa}, and subsequent pressure changes followed: \SI{20}{kPa} at \SI{0.7}{s}, \SI{30}{kPa} at \SI{1.4}{s}, \SI{40}{kPa} at \SI{2.3}{s}, and \SI{50}{kPa} at \SI{3.4}{s}. When the joint actuator reached \SI{50}{kPa}, the bottom actuator pressure slightly increased to \SI{18}{kPa}, likely due to the increased grasping force of the joint actuator.}
\add{When both the pump and the bottom actuator contributed, joint actuator pressure increased at \SI{20}{kPa/s}, twice the rate of \SI{10}{kPa/s} when only the pump was active.} 
\add{During perching, since the bottom inflatable actuator did not reach \SI{20}{kPa}, the pump was activated only when necessary to maintain the pressure at \SI{20}{kPa}.
Additionally, as shown in \figref{human_perching_30} A(P-2), even when the robot rotated \SI{1.7}{rad} in the roll direction, it is able to cling to the human arm securely.}

In the Deperch State, as illustrated in \figref{human_perching_30}B, immediately after takeoff,\add{the joint actuator pressure was \SI{17}{kPa}}.
The rest of the internal air of the joint inflatable actuator prevented the flexible arm from being fully stretched, resulting in \add{about \SI{0.4}{m} descent in the z direction} and unstable flight as illustrated in \figref{human_perching_30} A(D-2).
However, it could quickly recover balance in its posture.
\add{The remaining air was pushed out during the process of the arm's straightening by the thrust, and the joint actuator pressure eventually reached \SI{0}{kPa}.
The bottom actuator pressure dropped by about \SI{4}{kPa} during take-off due to the loss of grasping force. After returning to normal flight, the control returned to maintaining \SI{20}{kPa} as described in the approach state.}


\end{multicols}
\end{document}